\documentclass[11pt]{article}

\usepackage{acl}

\usepackage{geometry}
\usepackage{xcolor}
\usepackage{makecell}
\usepackage{times}
\usepackage{subcaption}
\usepackage{tcolorbox}
\usepackage{latexsym}
\usepackage{tabularx}
\usepackage{tabularray}
\usepackage{multicol}
\usepackage{multirow}%

\usepackage[T1]{fontenc}
\usepackage[utf8]{inputenc}
\usepackage{microtype}

\usepackage{hyperref}
\usepackage{subcaption}
\usepackage{graphicx}%
\usepackage{multirow}%
\usepackage{amssymb,amsfonts}%

\usepackage{amsthm}
\usepackage{mathrsfs}%
\usepackage[title]{appendix}%
\usepackage{textcomp}%
\usepackage{manyfoot}%
\usepackage{booktabs}%
\usepackage{natbib}
\usepackage{algorithmicx}%
\usepackage{algpseudocode}%
\usepackage{listings}%
\usepackage{CJKutf8}
\usepackage{float}
\usepackage{url}
\usepackage{tabularray}
\usepackage{soul}
\usepackage{color}
 \usepackage{colortbl}
 \usepackage{caption}
 \usepackage{array}
 \usepackage{tabularx}
 \usepackage{multicol}
 \usepackage[figuresright]{rotating}
 \usepackage[normalem]{ulem}
 
 \usepackage{geometry}
 \usepackage{multirow} 
 \usepackage{natbib}
 \usepackage{pifont}
 \DeclareCaptionLabelSeparator{colon}{. }

\usepackage{mathrsfs}%

\makeatletter
\newif\if@restonecol
\makeatother

\usepackage[linesnumbered,ruled,vlined]{algorithm2e}
\usepackage{algpseudocode}
\usepackage{amsmath}

\newcommand\blfootnote[1]{%
  \begingroup
  \renewcommand\thefootnote{}
  \footnotetext{#1}%
  \addtocounter{footnote}{-1}
  \endgroup
}

\title{
Linguistic Neuron Overlap Patterns
to Facilitate Cross-lingual Transfer
on Low-resource Languages
}

\author{Yuemei Xu$^{1}$, Kexin Xu$^{1}$, Jian Zhou$^{1}$, Ling Hu$^{1}$, Lin Gui$^{2}$\\
$^{1}$ School of Information Science and Technology, Beijing Foreign Studies University 
\\  $^{2}$  Department of Informatics, King's College London
\\
\{xuyuemei, xukexin, bwzj, huling\}@bfsu.edu.cn
\\  lin.1.gui@kcl.ac.uk 
}

\begin{document}
\maketitle
\begin{abstract}
The current Large Language Models (LLMs) 
face significant challenges in improving their performance on low-resource languages
and urgently need data-efficient methods without costly fine-tuning.
From the perspective of language-bridge,
we propose a simple yet effective method, namely BridgeX-ICL, 
to improve the zero-shot Cross-lingual In-Context Learning (X-ICL) for low-resource languages. 
Unlike existing works focusing on 
language-specific neurons,
BridgeX-ICL explores whether sharing
neurons can improve cross-lingual performance in LLMs.
We construct neuron probe data from the ground-truth MUSE 
bilingual dictionaries, and define a subset of language overlap neurons accordingly to ensure full activation of these anchored neurons.
Subsequently, we propose an HSIC-based metric to quantify LLMs' internal linguistic spectrum
based on overlapping neurons, 
guiding optimal bridge selection.
The experiments conducted on $4$ cross-lingual tasks and 
$15$ language pairs from $7$
diverse families, covering both high-low and moderate-low pairs, 
validate the effectiveness of BridgeX-ICL and offer empirical insights into the underlying multilingual mechanisms of LLMs. The code is publicly available at \url{https://github.com/xuyuemei/BridgeX-ICL}.
\end{abstract}

\section{Introduction}
\blfootnote{This work was supported by the National Social Science Foundation (No.24CYY107), the Fundamental Research Funds for the Central Universities (No.2024TD001), and the National Natural Science Foundation of China (No. 62576120).}
Although Large Language Models (LLMs) have demonstrated impressive multilingual capacities, there is still significant space for improving the performance on low-resource languages \cite{a22,a19}. To address this issue, especially avoiding costly post-training \cite{a23,a26}, it is critical to fully investigate the multilingual understanding and transferring ability in LLMs.

Recent research has increasingly focused on data-efficient methods, particularly Cross-lingual In-Context Learning (X-ICL) \cite{a27,a28,a19,a29},
which surprisingly works well on low-resource languages,
likely because LLMs are in-context low-resource language learners \cite{a11,a29}.
For instance, 
in the Arabic-to-Hebrew Bilingual Lexicon Induction (BLI) task, the zero-shot baseline accuracy in LLaMA 3 is 47.0\%. 
However,
simply specifying English as a bridge language in a zero-shot setting boosts accuracy to 64.5\%, 
which even significantly outperforms the two-shot X-ICL. 
This observation motivates us to further explore:
\textit{How can we improve cross-lingual capabilities of LLMs on low-resource languages by selecting an optimal bridge language in X-ICL}? 
Should the selection be purely data-driven, favoring high-resource bridge languages \cite{a21}? Or can human linguistic knowledge, such as language genealogy, or established evolutionary taxonomies, offer a more effective alternative \cite{a40,a30}?

To systematically investigate this issue, we leverage linguistic neurons \cite{a15} that handle language features to guide optimal bridge language selection in X-ICL.
However, there are two limitations when applying neuron-based interpretation \cite{a31, a15, a16} on low-resource languages:\\
\noindent $\bullet$ 
\textbf{Inaccurate neuron activation.} 
Current work often relies on multilingual corpora like Wikipedia \cite{a58} to probe internal neurons, without verifying whether LLMs truly understand the multilingual input.
This may lead to \textbf{unreliable neuron activations}, 
particularly for low-resource languages.
When LLMs poorly understand the probe input, they may instead activate neurons for processing unfamiliar or noisy input.
\\
\noindent $\bullet$ 
\textbf{Lacking guidance for cross-lingual transfer.}
Recent work argues that language-specific neurons do not facilitate cross-lingual transfer \cite{a50}.
This raises a critical question:
if language-specific neurons cannot, can sharing neurons improve cross-lingual transfer in LLMs? This exploration is also important for transferring language neuron research to actionable strategies to enhance the multilinguality of LLMs.

Motivated by this, we propose a simple yet effective bridge method, BridgeX-ICL, to improve LLMs' cross-lingual capabilities, especially on low-resource languages.
To address the \textbf{inaccurate activation issue}, we construct probe data by
leveraging the ground-truth bilingual lexicon MUSE \cite{x1}.
We collect bilingual word pairs
from MUSE that LLMs can translate accurately
and use them to prompt the models for bidirectional translations, generating answers in both language directions.
To address the \textbf{cross-lingual guidance issue},
we first explore overlap neurons' features
and their impact on cross-lingual transfer,
and then propose a bridge selection 
strategy based on the Hilbert-Schmidt Independence Criterion (HSIC) \cite{a59}.
Furthermore,
we measure the linguistic spectrum in LLMs based on overlapping-neurons and compare it with human language genealogy from Glottolog Trees \cite{a52}.
We conduct extensive 
experiments on $4$ cross-lingual tasks and $15$ language pairs from $7$ diverse families.
Our main contributions and findings are as 
follows:
\begin{itemize}
\item 
To the best of our knowledge,
this is the first work to explore 
language-bridge for zero-shot X-ICL
to improve LLMs' performance on low-resource languages.

\item We construct accurate neuron probe data 
and use it to fully activate
the anchored overlap neurons.
We also propose an HSIC-based metric to quantify the similarity 
between overlapping neurons and specific neurons for optimal bridge selection in X-ICL.

\item We validate the generalization of BridgeX-ICL on 
$4$ cross-lingual tasks and 
$15$ language pairs.
Here are empirical findings: 
1) Strong neural overlaps 
align with human linguistic taxonomy within
language families, but do not consistently hold across families.
2) Overlapping neurons
embody shared semantic information, regardless of the language within or between families.
3) BridgeX-ICL improves 
the performance on cross-lingual
tasks of BLI and MRC across 15 language pairs 
by an average of $6.02\%$ and $5.25\%$ over zero-shot baselines.
4) English is selected as the optimal bridge in 9 out of 15
language pairs in LLaMA (7 out of 15 in Mistral), indicating that high-resource, Latin-script languages tend to be the default
bridge. We also find that non-Latin script languages like Chinese also show potential as effective bridges.

\end{itemize}

\begin{figure*}[]
\centering
\includegraphics[width=0.93\textwidth]{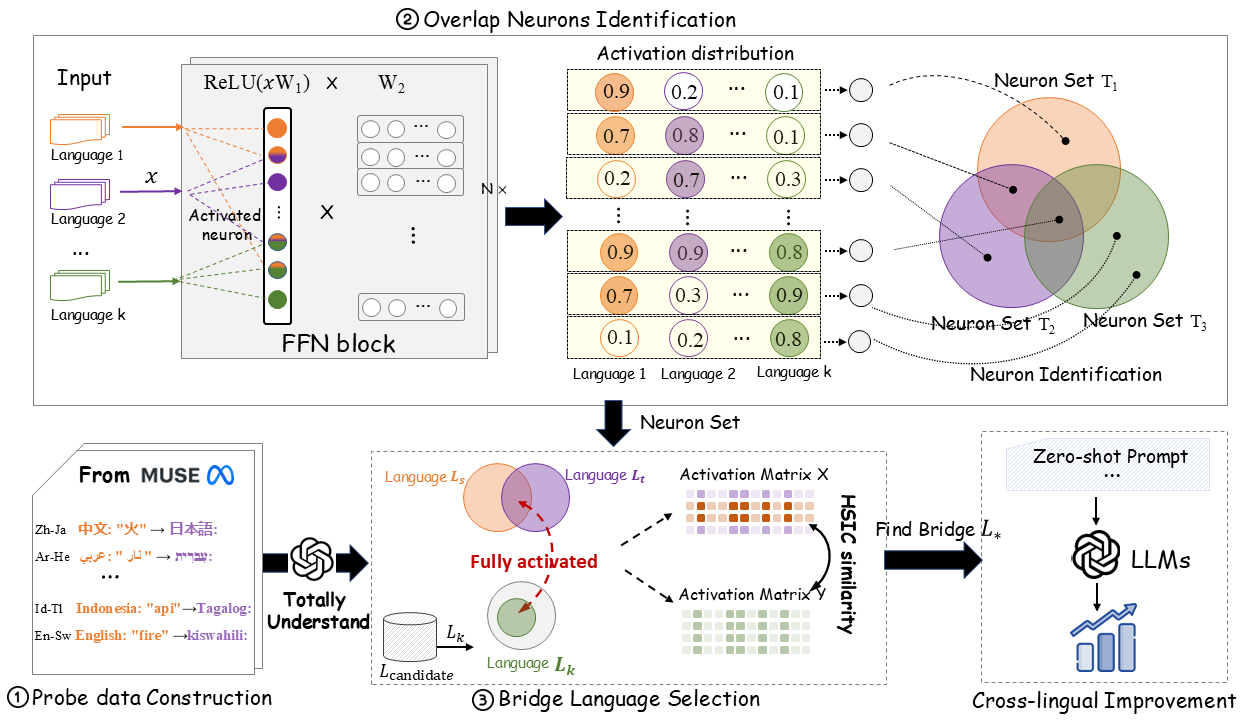}     
\caption{
\textbf{An illustration of BridgeX-ICL approach},
consisting of three steps:
Neuron probe data construction; 
Language neurons and their overlappings detection;
Optimal bridge $L_*$ selection
based on HSIC similarity.
}
\label{framework}
\end{figure*}

\section{Related Work}

\subsection{Cross-lingual In-context Learning}
LLMs face significant challenges when applied to low-resource languages \cite{a35, a38, a22}, mainly due to insufficient training data and 
\textit{the curse of multilinguality} \cite{a4}. To address these issues without updating model parameters, Cross-lingual In-context Learning (X-ICL), an extension of in-context learning (ICL), has recently gained attention \cite{a32}. Prior studies \cite{a27,a28,a19,a29} have demonstrated that LLMs act as effective few-shot multilingual learners, 
with few-shot ICL even outperforming fine-tuned language-specific models on several tasks \cite{a27}. However, 
few-shot X-ICL's performance is highly dependent on the context and the selection of examples, especially for unconventional or ambiguous languages \cite{a13,a19}. Consequently, existing research mainly focuses on optimizing few-shot example selection. To the best of our knowledge, we are the first to explore X-ICL explicitly from the perspective of leveraging language bridges.

\subsection{Linguistic Neuron in LLMs}
Recent research \cite{a40,a15,a16,a31,a30} has revealed that language-related neurons exist in FFN layers of transformer architecture. Deactivating these neurons will have a vital impact on LLMs' multilingual capacities. Beyond uncovering multilingual mechanisms, some research has gone to explore the neuron pattern across languages \cite{a30, a40} and its impact on cross-lingual performance \cite{a50,a61}. Specifically, Wang et al. (\citeyear{a30}) observed that similar languages may not exhibit significant neuron sharing in LLMs like BLOOM, suggesting that neuron sharing does not fully align with language similarity. Furthermore, recent work argues that language-specific neurons do not facilitate cross-lingual transfer \cite{a50}. This raises a critical question: whether sharing neurons can improve cross-lingual performance in LLMs. Motivated by these findings, we aim to further investigate LLM-internal neuron sharing across languages and its impact. In particular, we define a subset of language-overlapping neurons and explore whether they can serve as internal bridges to support cross-lingual inference.

\section{Methodology}
\subsection{Task Statement}

Given a set of languages $\mathcal{L}=\{L_1,...L_{|\mathcal{L}|}\}$,
this work aims to measure 
the linguistic genealogy implicitly learned by LLMs from language overlapping neurons, then use the quantified linguistic similarity to 
guide the bridge language selection in X-ICL.

Figure \ref{framework} depicts three main steps of our approach:
\ding{172} neuron probe data construction;
\ding{173} language neurons and their overlapping detection;
\ding{174} bridge language selection,
guided by the observed pattern of 
overlapping neurons and a modified HSIC dependency estimation, 
which
selects the optimal $L_*$ from the candidate set $\mathcal{L}_\text{candidate}$ to facilitate X-ICL from a source language $L_s$ to a target language $L_t$.

\subsection{Probe Data Construction}
We employ two types of probe data
for different purposes of 
language neurons identification and optimal bridge selection.
The former is task independent, targeting language neurons,
and can use existing multilingual corpora.
In our work, we adopt FLORES+ \cite{x7}, a high-quality parallel corpus released by Meta,
and combine its development set and test set to obtain $2,000$ parallel sentences for each language.

Bridge selection for X-ICL 
needs to consider both language neurons and those that contribute to cross-lingual tasks.
Inspired by findings on task-specific neurons \cite{a57},
we propose that certain neurons directly influence cross-lingual transfer, 
and their manipulation and measurement
should not rely solely on monolingual corpora.
Therefore, we construct probe data
by leveraging bilingual word translations as follows.

\textbf{Probe Data Design.}
We collect $d$ (i.e., $100$) word pairs that LLMs can accurately translate.
These word pairs are fed into the LLMs in both directions of $L_1 \to L_2$ and $L_2\to L_1$,
ensuring neurons linked to $L_1$ and $L_2$
are fully activated. 
Instead of feeding word pairs directly, we 
prompt LLMs to generate translations, 
which guarantees accurate neuron activation. 
Examples of probe data for $3$ language pairs are shown below.

\begin{figure}[!h]
\centering
\includegraphics[width=0.4\textwidth]{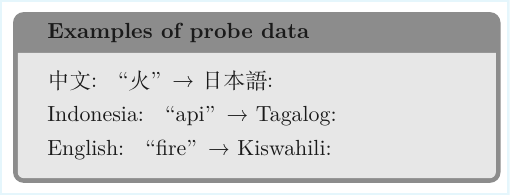}
\end{figure}

\subsection{Linguistic Overlap Neurons}
\subsubsection{Neurons in LLMs}

Neuron identification follows \cite{a15},
which assumes that language neurons are mainly located 
in the Feed-Forward Network (FFN) layers.
Given the transformation at the $i$-th layer:
\begin{equation}
\boldsymbol{h}_i = 
\sigma (\tilde{\boldsymbol{h}_i} \boldsymbol{W}_1^i ) \cdot \boldsymbol{W}_2^i
\end{equation}
where $\tilde{\boldsymbol{h}_i}$ is
the hidden state input to the $i$-th layer 
and $\sigma(\cdot)$ denotes the activation function.
$\boldsymbol{W}_1^i \in \mathbb{R}^{d\times N}$ and 
$\boldsymbol{W}_2^i \in \mathbb{R}^{N \times d}$ 
are the learned parameters.
Here, a neuron is defined as 
a linear transformation of a single column in
$\boldsymbol{W}_1^i$ and there are $N$ neurons
in each layer.
The activation value of the $j$-th neuron is $\sigma(\tilde{\boldsymbol{h}_i} \boldsymbol{W}_1^i)_j$.
If this value exceeds 0, the neuron is considered an activated neuron.

\subsubsection{Overlap Neuron Identification}
\label{sec.overlapi}

First, we identify neurons $\mathcal{T}_k$ associated with each language $L_k$.
Unlike existing work \cite{a30,a50} using
\textbf{L}anguage \textbf{A}ctivation \textbf{P}robability \textbf{E}ntropy (LAPE) \cite{a15} to identify 
neurons with high activation probability for one language 
but low for others,
which is less effective to capture
neuron relationships
across languages,
we identify neurons set $\mathcal{T}_k$ based on activation frequency.
Let $f_{k,j}$ denote the activation frequency of neuron $n_j$ 
when processing tokens from language $L_k$.
Neurons with the top $\tau \cdot N$
activation frequencies are selected into 
$\mathcal{T}_k$ based on 
a threshold $\tau$.

\noindent
\textbf{Overlap Neuron Definition.}
For languages $L_u$ and $L_v$, 
with associated neuron sets 
$\mathcal{T}_u$ and $\mathcal{T}_v$,
the overlap neurons are defined as the interaction of $\mathcal{T}_u$ and $\mathcal{T}_v$.
At the $i$-th FFN layer, we have $\mathcal{T}_{u,v}(i)=\mathcal{T}_u(i) \cap \mathcal{T}_v(i)$.

\noindent
\textbf{Linguistic Similarity Calculation.}
The linguistic similarity between $L_u$ and $L_v$ is quantified
through the activation frequencies of their overlapping
neurons.
Let $\boldsymbol{f_u}=\{f_{u,1},f_{u,2},...,f_{u,|\mathcal{T}_{u,v}|}\}$
denote the activation frequency vector of neurons
in $\mathcal{T}_{u,v}$ when processing 
tokens from $L_u$.
The linguistic similarity between $L_u$ and $L_v$ is calculated:
\begin{equation}
\text{sim}(\mathcal{T}_u,\mathcal{T}_v) = \frac{\boldsymbol{f_u} \cdot \boldsymbol{f_v}}{\lVert \boldsymbol{f_u} \rVert \lVert \boldsymbol{f_v} \rVert}
\label{eq:sim}
\end{equation}
By computing pairwise similarities for all languages in $\mathcal{L}$,
we obtain a comprehensive linguistic spectrum of the model.

\subsubsection{Overlap Neuron Pattern}
\label{neuronovs}
Second, we use the constructed probe data from FLORES+
to explore overlapping neurons' features and their 
generalized impact on cross-lingual transfer
so that we can utilize them to 
guide bridge 
language selection.
We have two observations:

\begin{figure}[!h]
\centering
\includegraphics[width=0.48\textwidth]{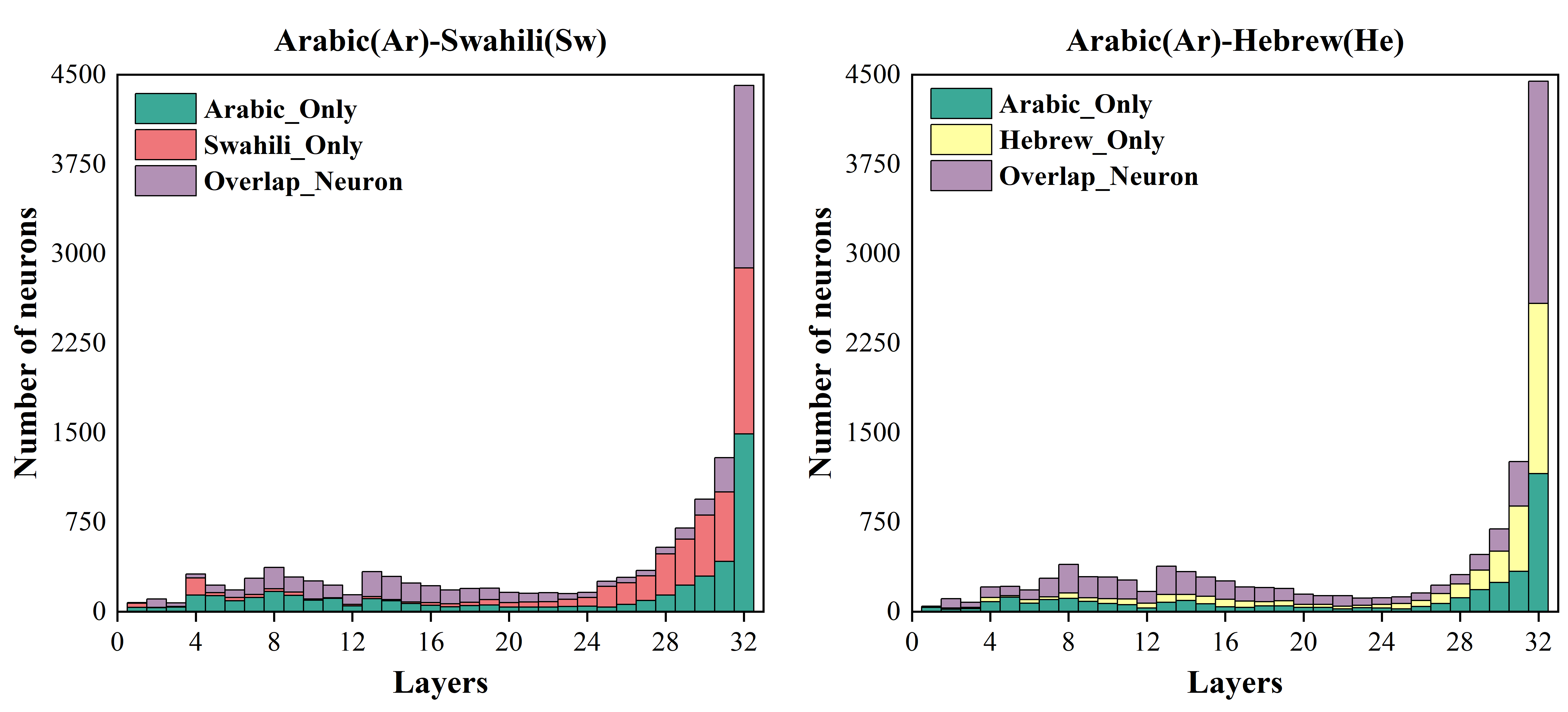}  
\caption{
\textbf{Language-overlapping neurons} on distant pair (Arabic-Swahili) and close pair (Arabic-Hebrew).
}
\label{fig:vis-overlap}
\end{figure}

\begin{figure}[!ht]
\centering
\includegraphics[width=0.47\textwidth]{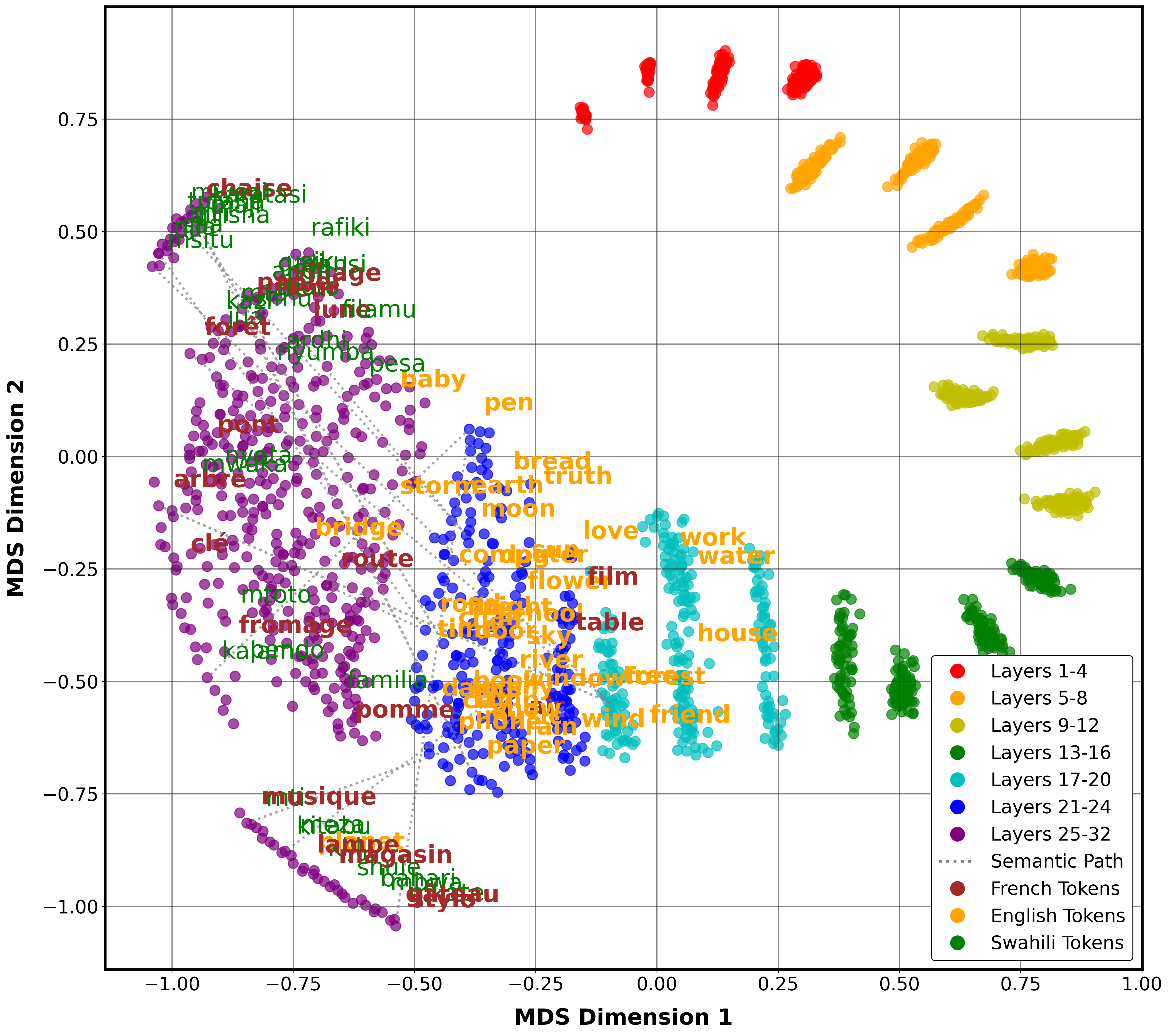}   
\caption{
\textbf{Layer-wise latent embeddings projected with MDS}
in French-Swahili translation.
A rainbow-colored path traces the latent embeddings across 32 layers.
The predicted Swahili tokens are in green, and their correct English tokens are in orange.}
\label{fig:pathvisuall}
\end{figure}

\noindent $\bullet$ 
Similar languages 
share more neurons than 
distant ones.
For example,
Arabic-Hebrew within the same language family 
has more overlapping neurons than 
Arabic-Swahili across families, 
as presented in Figure \ref{fig:vis-overlap}.
This suggests the potential of neural overlap to
measure language distance.

\noindent $\bullet$ 
Overlap neurons are
predominantly concentrated in the middle and 
final layers,
serving distinct roles of semantic understanding and language decoding
for next-token prediction.
This neural function is further evidenced by 
neurons deactivation shown in Figure \ref{fig8:mistddd}.
Specifically,
neurons in final layers are task-related and responsible for cross-lingual generation.
To examine whether middle-layer neurons handle semantic understanding,
we further employ a technique called \textit{logit lens} \cite{a60} 
to visualize the latent semantic embeddings across layers.
We visualize French-Swahili translation 
using $60$ word pairs that LLaMA 3 translates accurately.
We extract the model's latent embeddings 
at each layer for next-token prediction
and project them into a 2D space using 
classical multidimensional scaling (MDS), 
presented in Figure \ref{fig:pathvisuall}.
The embedding trajectory is marked in a rainbow-colored path
(e.g., red = layers 1-4, violet = layers 25-32).
We can observe French inputs and 
their corresponding correct English next
tokens cluster in middle layers,
indicating that LLMs rely on the knowledge in high-resource languages like English to perform cross-lingual reasoning there.
Neurons in middle layers
should be prioritized over 
those in final layers when quantifying language similarity.

\subsection{Bridge Language Selection}
Based on the above observations, 
we leverage the constructed probe data with $d$ samples per language pair to identify the 
optimal bridge language $L_*$ to facilitate X-ICL from source language $L_s$ to target language $L_t$.

Given $L_s$ and $L_t$ and their overlap neurons $\mathcal{T}_{s,t}$ identified in section \ref{sec.overlapi},
we obtain the activation value matrix $\mathbf{X} \in \mathbb{R}^{|\mathcal{T}_{s,t}|\times 2d}$ by prompting the LLM with $d$ samples in both directions
for balanced neuron activation in $L_s$ and $L_t$.
We also obtain 
the activation matrix
$\mathbf{Y} \in \mathbb{R}^{|\bar{\mathcal{T}_{y}|}\times 2d}$ for a candidate bridge language $L_y \in \mathcal{L}_\text{candidate}$.
Here,
$\bar{\mathcal{T}_{y}}=\mathcal{T}_{y}- \mathcal{T}_{s,t}-\mathcal{T}_{y'}$ represents
the set of 
language-specific neurons in $L_y$,
excluding those shared with $\mathcal{T}_{s,t}$ or 
with any other language $L_{y'}\neq L_y$.

We employ HSIC \cite{a59} to
measure the nonlinear dependency between activation matrices $\mathbf{X}$
and $\mathbf{Y}$.
Average pooling will be performed to standardize matrices of $\mathbf{X}$ and $\mathbf{Y}$ to have the same row dimension $n$.
The HSIC is formally calculated as:
$\text{HSIC}(\mathbf{X},\mathbf{Y}) = n^{-2}\text{Tr}(\mathbf{KHLH})
$,
where $\mathrm{Tr}(\cdot)$ is the trace operation, 
$\mathbf{K},\mathbf{L} \in \mathbb{R}^{n \times n}$ are learned kernel matrices for $\mathbf{X}$
and $\mathbf{Y}$.
$\mathbf{H}=\mathbf{I}_{n\times n}- \frac{1}{n} \mathbf{1}_n \mathbf{1}_n^\top$ is a centering matrix, where
$\mathbf{I}_{n\times n}$ is the identity matrix of size $n\times n$, $\mathbf{1}_n$ is a vector of $n$ ones.
Rather than computing HSIC over the entire activation matrices, we adopt a bidirectional maximum matching strategy that
measures the strongest dependency between 
a single neuron vector $\mathbf{x}_i \in \mathbf{X}$ ($\mathbf{y}_j \in \mathbf{Y}$)
and the entire distribution of the other,
where $\mathbf{x}_i, \mathbf{y}_i\in \mathbb{R}^d$, computed as:

\begin{small}
\begin{equation}
\mathrm{H}(\bar{\mathcal{T}}_{y},\mathcal{T}_{s,t})
= \frac{1}{2} \Big(
   \max_{i}\,\mathrm{HSIC}(\mathbf{x}_i,\mathbf{Y})
   + \max_{j}\,\mathrm{HSIC}(\mathbf{X},\mathbf{y}_j)
 \Big)
\end{equation}
\end{small}
We compute the dependency scores layer by layer
and average them across the middle $K$ layers to 
estimate the selection probability of $L_y$:
\begin{equation}
\begin{aligned}
p(L_y|L_s \rightarrow L_t)=\frac{1}{K}\sum_{i=1}^K \text{H}\left( \bar{\mathcal{T}}_{y}(i),\mathcal{T}_{s,t}(i) \right)
\label{equationss}
\end{aligned}
\end{equation}
where $K$ is determined according
to embedding semantic similarity and discussed in section \ref{sec:appendixB_k_dis}.
Finally, the optimal bridge $L^*$ is selected by:
\begin{equation}
L_* = \arg\max_{L_y \in \mathcal{L}_\text{candidate}}p(L_y|L_s \rightarrow L_t)
\label{eq:optimal}
\end{equation}

\section{Experiment}
\subsection{Experiment Setup}
\noindent \textbf{Implementation.}
We evaluate BridgeX-ICL on $4$ cross-lingual tasks and $15$ languages covering
$7$ diverse language families: 
\textbf{Indo-European}: 
English (En), German (De), French (Fr),
Italian (It),
Portuguese (Pt), Spanish (Es);
\textbf{Uralic}: 
Finnish (Fi), Hungarian (Hu);
\textbf{Afro-Asiatic}: 
Arabic (Ar), Hebrew (He);
\textbf{Austronesian}:
Indonesian (Id), Tagalog (Tl);
\textbf{Sino-Tibetan}: Chinese (Zh);
\textbf{Japonic}: Japanese (Ja);
\textbf{Niger-Congo}: Swahili (Sw).

As the evaluation focuses on LLMs' cross-lingual transfer on low-resource languages,
we take 
He, Tl, Sw, and Ja as target languages
to build $15$ cross-lingual pairs,
covering moderate-to-low (e.g., Ar-Sw) and high-to-low (e.g., En-He) pairs, both within and across language families.
The classification of high-, moderate-, and low-resource languages is based on their proportion in LLMs' training corpora, following previous work \cite{a34}.
Since the bridge language should be 
well supported by LLMs,
we select $6$ languages in the Indo-European family as candidate bridges. 
We also
conducted an exploratory experiment using
bridges not in the Indo-European family
discussed in
section \ref{dis:candidate}.

\vspace{7pt}
\noindent \textbf{Datasets.} 
To evaluate the generalization of bridge selection beyond the BLI task,
we further evaluate cross-lingual 
Machine Reading Comprehension (MRC) using the Belebele dataset \cite{x11}.
To verify robustness,
we additionally consider two cross-lingual tasks:
Cross-Lingual Question Answering (CLQA)
and Cross-Lingual Natural Language Inference (XNLI).
Since CLQA and XNLI 
cover only six of our evaluated language pairs,
their experimental results are reported in Appendix \ref{sec: task}.

To evaluate low-resource languages,
a key challenge lies in the lack of evaluation benchmarks.
Although the ground-truth MUSE \cite{x1} provides $110$ bilingual dictionaries for the BLI task,
it does not cover the $15$ language pairs we tested.
To address this, 
we used English as a pivot to build $L_s$-$L_t$ dictionary
from $L_s$-$\text{English}$ and $L_t$-$\text{English}$.
For languages not in MUSE (e.g., Swahili), we extracted word pairs from wiktionary\_bli \cite{x2} to build En-Sw. We verified all word pairs using both Google and Microsoft translators to ensure quality and selected $1,000$ word pairs for each language pair that are consistently validated by both systems \footnotemark[1].
\footnotetext[1]{The constructed BLI dictionaries are available at: \url{https://github.com/xuyuemei/BLI-}.}

\vspace{10pt}
\noindent \textbf{Metrics.}
For the BLI task,
we use the Precision@N metric, which measures the accuracy of the model's 
top-$N$ candidate translations.
In this study, $N$ is set to $1$. For the MRC task, we use accuracy to evaluate
whether the model selects the correct answer from multiple choices.

\vspace{10pt}
\noindent \textbf{LLMs.}
We conducted experiments on two open-source LLMs: LLaMA-3-8B \cite{a3}  and Mistral-7B-Instruct-v0.3 \cite{a45}.
Their training corpora cover 176 and 53 languages, respectively, 
which include all the experimental low-resource languages and allow us to explore the underlying linguistic mechanisms.

\vspace{10pt}
\noindent \textbf{Baselines.}
Baselines are divided into \textit{zero-shot},
\textit{few-shot}, and \textit{zero-shot with bridge}.
Specifically,
\textit{zero-shot} is the basic prompt setup, and \textit{few-shot} 
builds on the zero-shot prompt by adding
1, 2, 3, or 4 samples.
For zero-shot with bridge approach,
we compare BridgeX-ICL with 
5 baselines described below.
1) \textbf{Phylogenetic Distance Source/Target} (Ph.D Source or Ph.D Target): 
Select the bridge language closest to the source or target language according to
language genealogy of Glottolog Trees \cite{a52}.
2) \textbf{English Bridge}:
Use English as the bridge language.
3) \textbf{Sharing Matters}: 
\citeauthor{a30}(\citeyear{a30}) used activation values to find shared neurons across languages. We select language with the most shared neurons as the bridge language.
4) \textbf{IoU}: Use Intersection over Union (IoU) \cite{x12}, also known as Jaccard index, to measure linguistic distance. 
Given neuron sets $\mathcal{T}_u, \mathcal{T}_v$
associated with language $L_u$, $L_v$, 
$\mathrm{IoU}(\mathcal{T}_{u},\mathcal{T}_{v}) = |\mathcal{T}_{u}\cap\mathcal{T}_{v}| / |\mathcal{T}_{u}\cup\mathcal{T}_{v}|$.
Language with the highest average IoU score to $L_s$ and $L_t$ is selected.
5) 
\textbf{LAPE\_{\text{overlap}}}: Use entropy-based LAPE 
\cite{a15} to identify language-specific neurons. We then compute cosine similarity on overlap neurons between the bridge and source/target. The language with the highest average similarity is selected.

\begin{figure}[!h] 
  \centering
  \begin{minipage}[b]{0.48\textwidth} 
    \centering
    \includegraphics[width=\linewidth, keepaspectratio]{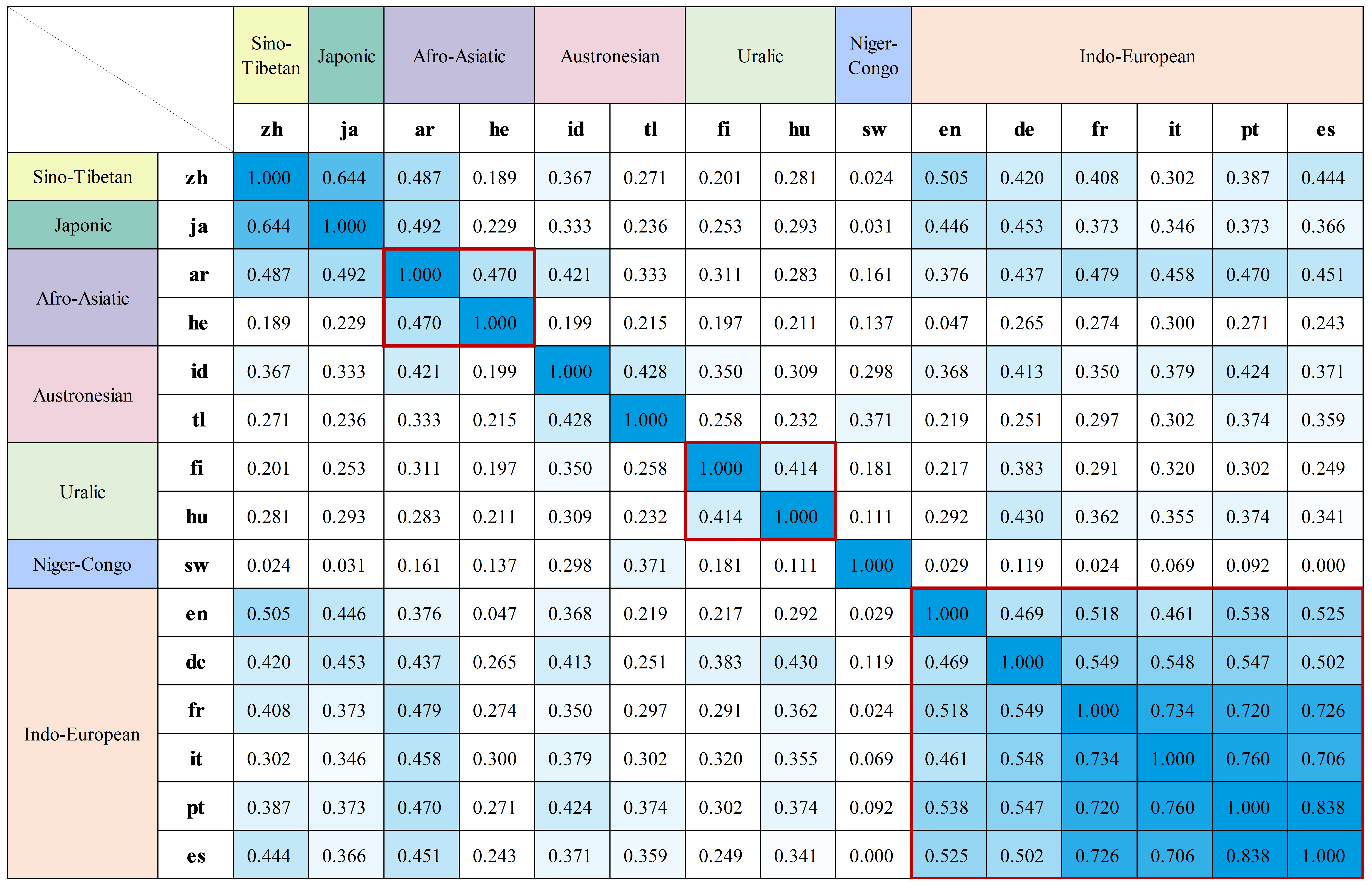} 
    (a) Linguistic spectrum in LLaMA 3
    \label{fig:subfigA}
  \end{minipage}
  \hfill 
  \begin{minipage}[b]{0.48\textwidth} 
    \centering
    \includegraphics[width=\linewidth, keepaspectratio]{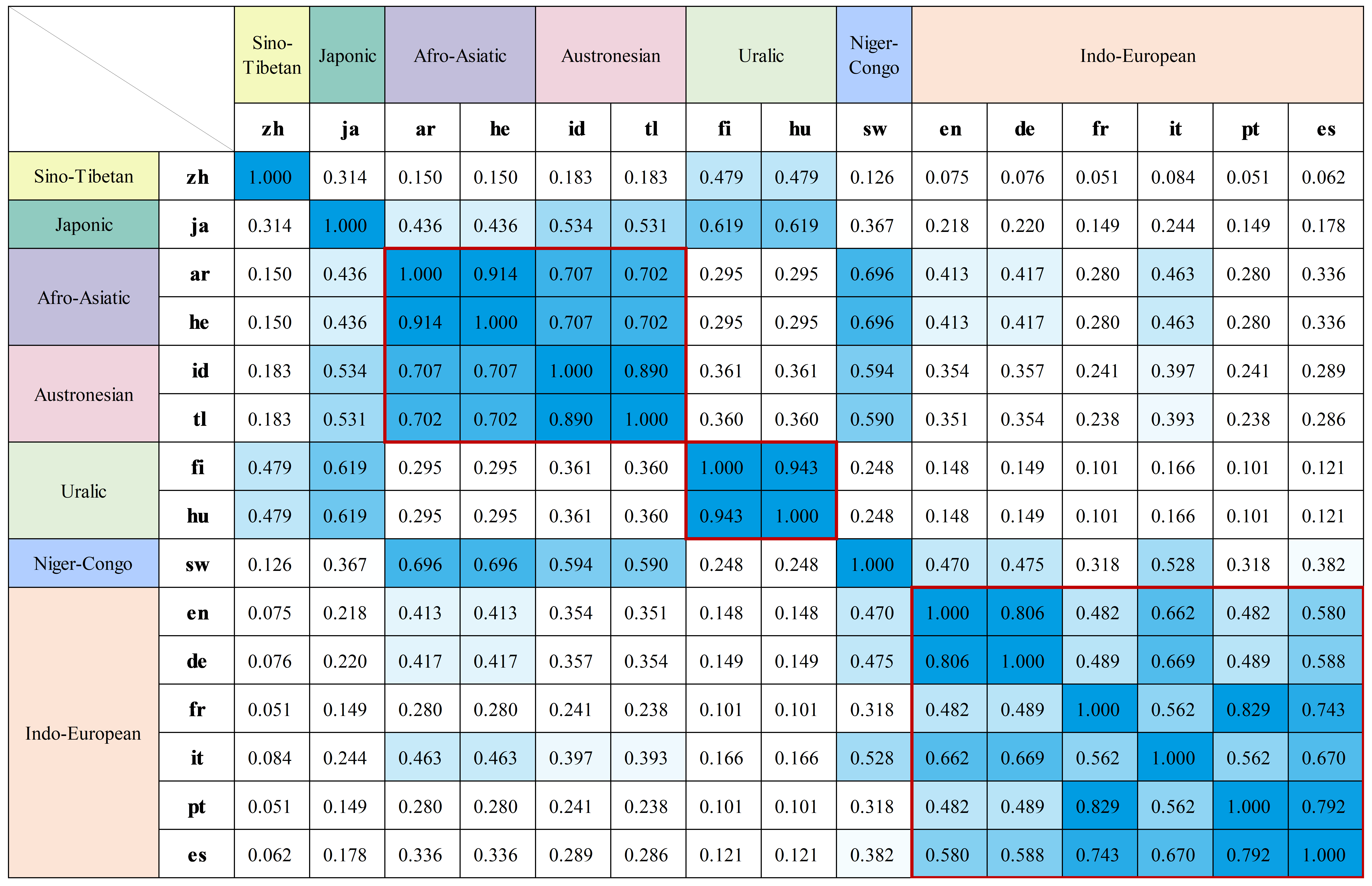} 
    (b) Language similarity from Glottolog Trees 
    \label{fig:subfigB}
  \end{minipage}
  \caption{
  \textbf{Comparison of linguistic spectrum} calculated based on overlapping neurons
  in LLaMA 3 and language similarity derived from Glottolog Phylogenetic Trees,
  including $15$ languages from $7$ families. 
  Darker blue indicates a higher language similarity.}
  \label{fig:heatmep}
\end{figure}

\subsection{Main Results}
\subsubsection{LLMs' Linguistic Spectrum Discussion}
This section discusses the linguistic similarities 
across $15$ languages from $7$ families,
calculated based on overlapping neurons in 
LLaMA 3 and Mistral,
as presented in Figure \ref{fig:heatmep}(a) and Figure \ref{fig:activemistralspr} in Appendix~\ref{sec:appendixB_Mistral}, respectively.
To evaluate how closely 
the linguistic spectrum learned by LLMs
aligns with that of human languages,
we leverage Glottolog Phylogenetic Trees \cite{a52}, which encode hierarchical relationships among 8,000+ human languages,
to derive human language similarity in Figure \ref{fig:heatmep}(b).
The detailed process for computing linguistic similarity based on Glottolog Trees is in Appendix~\ref{sec:appendixA}.
In Figure \ref{fig:heatmep},
darker blue indicates stronger similarity,
and the diagonal denotes self-similarity (1.0).

\textbf{Linguistic spectrum learned by LLMs is not fully aligned with 
human language phylogeny}.
We observe
strong neural similarities within language families, 
highlighted by a red text box in Figure \ref{fig:heatmep},
which matches human linguistic taxonomy.
For example, 
Arabic (Ar) and Hebrew (He),
within the Afro-Asiatic family,
exhibit a high neuron overlap (0.470), greater than Arabic-Swahili with 0.161 similarity.
A similar pattern appears with Indonesian (Id) and Tagalog (Tl), both from the Austronesian family.
In addition,
high-resource Indo-European languages, 
such as 
French-Italian (Fr-It) and Portuguese-Spanish (Pt-Es), 
show the highest overlap scores,
with the darkest blue in the bottom-right corner of the heatmap.
But this alignment breaks down in high-to-low resource language pairs, and some cross-family pairs display unexpected high similarity scores, likely reflecting training data distribution rather than intrinsic linguistic relationships.

\textbf{LLMs build their own distinct understanding of language relationships.}
The calculated linguistic spectra of LLaMA 3 and Mistral 
are similar but not identical.
The two models may choose different bridges for the same language pair, 
as discussed later.
We observe that Arabic has the strongest similarity (0.479) with French in the Romance family, rather than with Hebrew (0.470) from its own Afro-Asiatic family.
This counterintuitive result
is likely due to 
the linguistic relationships
learned by LLMs
being primarily shaped by the distribution of languages in training corpora, as noted in \cite{a13}.

\begin{table*}[!htb]
\caption{Comparison of BLI task improvement on $15$ language pairs. The highest gains are marked with \textbf{bold} in
few-shot and zero-shot with bridge methods.
`-'  indicates the selected bridge is either the source or target language.}
\label{table:BLI}
\resizebox{\textwidth}{!}{
\centering
\renewcommand{\arraystretch}{1.1}\begin{tabular}
{clllllllllllllllll}
\toprule[1pt]
\multicolumn{17}{c}{\Large LLaMA-3-8B} \rule{0pt}{13pt} \\ \hline
\multicolumn{2}{c|}{\textbf{Method}} &
  \multicolumn{1}{c}{{\textbf{Zh-Ja}}} &
  \multicolumn{1}{c}{{\textbf{Zh-He}}} &
  \multicolumn{1}{c}{{\textbf{Zh-Tl}}} &
  \multicolumn{1}{c}{{\textbf{Zh-Sw}}} &
  \multicolumn{1}{c}{{\textbf{Ar-Ja}}} &
  \multicolumn{1}{c}{{\textbf{Ar-He}}} &
  \multicolumn{1}{c}{{\textbf{Ar-Tl}}} &
  \multicolumn{1}{c}{{\textbf{Ar-Sw}}} &
  \multicolumn{1}{c}{{\textbf{Id-Ja}}} &
  \multicolumn{1}{c}{{\textbf{Id-He}}} &
  \multicolumn{1}{c}{\textbf{Id-Tl}} &
  \multicolumn{1}{c}{\textbf{Id-Sw}} &
  \multicolumn{1}{c}{\textbf{En-He}} &
  \multicolumn{1}{c}{\textbf{En-Tl}} &
  \multicolumn{1}{c}{\textbf{En-Sw}} \\ \hline
\multicolumn{2}{c|}{Zero-shot} &
  \multicolumn{1}{c}{67.10} &
  \multicolumn{1}{c}{44.10} &
  \multicolumn{1}{c}{42.60} &
  \multicolumn{1}{c}{31.20} &
  \multicolumn{1}{c}{69.90} &
  \multicolumn{1}{c}{47.00} &
  \multicolumn{1}{c}{46.70} &
  \multicolumn{1}{c}{39.10} &
  \multicolumn{1}{c}{62.50} &
  \multicolumn{1}{c}{44.70} &
  \multicolumn{1}{c}{49.30}&
  \multicolumn{1}{c}{25.90}&
  \multicolumn{1}{c}{56.90}&
\multicolumn{1}{c}{60.00}&
  \multicolumn{1}{c}{28.80} & \\ \hline
\multicolumn{1}{c|}{} &
  \multicolumn{1}{l|}{One-shot} &
  \multicolumn{1}{c}{+3.20} &
  \multicolumn{1}{c}{+12.60} &
  \multicolumn{1}{c}{+1.20} &
  \multicolumn{1}{c}{+3.00} &
  \multicolumn{1}{c}{+4.20} &
   \multicolumn{1}{c}{+13.50} &
   \multicolumn{1}{c}{-4.80} &
   \multicolumn{1}{c}{+0.60} &
   \multicolumn{1}{c}{+7.40} &
  \multicolumn{1}{c}{+7.50}  &
   \multicolumn{1}{c}{+0.70} &
   \multicolumn{1}{c}{+6.00}&
   \multicolumn{1}{c}{+18.60}&
   \multicolumn{1}{c}{-6.30}&
   \multicolumn{1}{c}{+4.40}&
   \\
\multicolumn{1}{c|}{} &
  \multicolumn{1}{l|}{Two-shot} &
  \multicolumn{1}{c}{+9.40} &
  \multicolumn{1}{c}{+16.90} &
  \multicolumn{1}{c}{+2.20} &
  \multicolumn{1}{c}{+5.10} &
  \multicolumn{1}{c}{\textbf{+9.40}} &
  \multicolumn{1}{c}{+13.90} &
   \multicolumn{1}{c}{-0.30} &
   \multicolumn{1}{c}{+3.50} &
  \multicolumn{1}{c}{+16.00} &
    \multicolumn{1}{c}{+15.90}&
   \multicolumn{1}{c}{+5.90} &
    \multicolumn{1}{c}{\textbf{+6.10}}&
   \multicolumn{1}{c}{+23.90} &
   \multicolumn{1}{c}{-5.50} &
   \multicolumn{1}{c}{+6.50} &
   \\
\multicolumn{1}{c|}{} &
  \multicolumn{1}{l|}{Three-shot} &
  \multicolumn{1}{c}{+6.70} &
  \multicolumn{1}{c}{\textbf{+22.20}} &
  \multicolumn{1}{c}{\textbf{+3.70}} &
  \multicolumn{1}{c}{+6.80} &
  \multicolumn{1}{c}{+7.80} &
   \multicolumn{1}{c}{\textbf{+16.90}}&
   \multicolumn{1}{c}{\textbf{+1.50}}&
   \multicolumn{1}{c}{\textbf{+3.70}}&
   \multicolumn{1}{c}{\textbf{+16.70}}&
   \multicolumn{1}{c}{\textbf{+22.90}}&
   \multicolumn{1}{c}{+10.10}&
   \multicolumn{1}{c}{-4.30}&
   \multicolumn{1}{c}{\textbf{+26.30}}&
   \multicolumn{1}{c}{-4.00}&
   \multicolumn{1}{c}{+6.90}&
   \\
\multicolumn{1}{c|}{\multirow{-4}{*}{Few-shot}} &
  \multicolumn{1}{l|}{Four-shot} &
  \multicolumn{1}{c}{\textbf{+12.50}} &
  \multicolumn{1}{c}{+20.50} &
  \multicolumn{1}{c}{+3.20} &
  \multicolumn{1}{c}{\textbf{+7.00}} &
  \multicolumn{1}{c}{+7.70} &
  \multicolumn{1}{c}{+15.80} &
   \multicolumn{1}{c}{+0.80}&
  \multicolumn{1}{c}{+3.20} &
   \multicolumn{1}{c}{+14.30}&
   \multicolumn{1}{c}{+22.00}&
   \multicolumn{1}{c}{\textbf{+10.60}}&
   \multicolumn{1}{c}{-6.00}&
   \multicolumn{1}{c}{+26.10} &
   \multicolumn{1}{c}{\textbf{-3.70}}&
  \multicolumn{1}{c}{\textbf{+7.40}}&
   \\ \hline
\multicolumn{1}{c|}{} &
  \multicolumn{1}{l|}{Ph.D Source} &
   \multicolumn{1}{c}{+2.80}&
   \multicolumn{1}{c}{+6.70}  &
   \multicolumn{1}{c}{+3.80} &
   \multicolumn{1}{c}{+1.50}&
   \multicolumn{1}{c}{-9.50}&
   \multicolumn{1}{c}{-}&
   \multicolumn{1}{c}{-7.40}&
   \multicolumn{1}{c}{-0.60}&
   \multicolumn{1}{c}{-11.90}&
   \multicolumn{1}{c}{-3.20}&
   \multicolumn{1}{c}{-}&
  \multicolumn{1}{c}{-3.90} &
   \multicolumn{1}{c}{+12.30}&
   \multicolumn{1}{c}{-3.30}&
   \multicolumn{1}{c}{-1.70}&
   \\
\multicolumn{1}{c|}{} &
  \multicolumn{1}{l|}{Ph.D Target} &
   \multicolumn{1}{c}{+2.80}&
   \multicolumn{1}{c}{+9.60}&
   \multicolumn{1}{c}{-0.10}&
   \multicolumn{1}{c}{\textbf{+5.00}}&
   \multicolumn{1}{c}{-0.50}&
   \multicolumn{1}{c}{-}&
   \multicolumn{1}{c}{-7.10}&
   \multicolumn{1}{c}{-}&
   \multicolumn{1}{c}{-12.10}&
   \multicolumn{1}{c}{+3.80}&
   \multicolumn{1}{c}{-}&
   \multicolumn{1}{c}{+1.20}&
   \multicolumn{1}{c}{\textbf{+16.40}}&
   \multicolumn{1}{c}{-2.30}&
   \multicolumn{1}{c}{\textbf{+2.30}}&
   \\
\multicolumn{1}{c|}{} &
  \multicolumn{1}{l|}{English Bridge} &
  \multicolumn{1}{c}{\textbf{+10.80}}  &
  \multicolumn{1}{c}{+12.60}  &
  \multicolumn{1}{c}{\textbf{+11.40}}  &
  \multicolumn{1}{c}{+4.80}&
  \multicolumn{1}{c}{\textbf{+10.70}} &
  \multicolumn{1}{c}{\textbf{+17.50}} &
   \multicolumn{1}{c}{\textbf{+10.10}} &
   \multicolumn{1}{c}{+3.30} &
   \multicolumn{1}{c}{\textbf{+3.60}} &
   \multicolumn{1}{c}{+9.30} &
   \multicolumn{1}{c}{+10.40}&
  \multicolumn{1}{c}{+2.50} &
  \multicolumn{1}{c}{-} &
   \multicolumn{1}{c}{-}&
   \multicolumn{1}{c}{-}&
   \\
\multicolumn{1}{c|}{} &
  \multicolumn{1}{l|}{Sharing Matters} &
   \multicolumn{1}{c}{+9.50} &
   \multicolumn{1}{c}{\textbf{+14.50}}&
   \multicolumn{1}{c}{+8.40}&
   \multicolumn{1}{c}{+2.60}&
   \multicolumn{1}{c}{+6.40}&
   \multicolumn{1}{c}{+17.10}&
   \multicolumn{1}{c}{+6.10}&
   \multicolumn{1}{c}{\textbf{+3.90}}&
   \multicolumn{1}{c}{+2.60}&
   \multicolumn{1}{c}{+15.20}&
   \multicolumn{1}{c}{+7.70}&
   \multicolumn{1}{c}{+2.40}&
   \multicolumn{1}{c}{+12.30}&
   \multicolumn{1}{c}{-3.30}&
   \multicolumn{1}{c}{-1.70}&
   \\
\multicolumn{1}{c|}{} &
  \multicolumn{1}{l|}{IoU Score} &
  \multicolumn{1}{c}{\textbf{+10.80}}  &
  \multicolumn{1}{c}{+12.60}  &
  \multicolumn{1}{c}{\textbf{+11.40}} &
  \multicolumn{1}{c}{+4.80}  &
  \multicolumn{1}{c}{\textbf{+10.70}}&
  \multicolumn{1}{c}{+13.20} &
  \multicolumn{1}{c}{+5.30} &
   \multicolumn{1}{c}{+3.50} &
   \multicolumn{1}{c}{+2.30} &
   \multicolumn{1}{c}{+15.20} &
   \multicolumn{1}{c}{+7.40}&
  \multicolumn{1}{c}{+3.80} &
  \multicolumn{1}{c}{+9.70} &
   \multicolumn{1}{c}{-6.10}&
   \multicolumn{1}{c}{-1.80}&
   \\
\multicolumn{1}{c|}{} &
  \multicolumn{1}{l|}{LAPE\_{\text{overlap}}} &
  \multicolumn{1}{c}{+10.50}  &
  \multicolumn{1}{c}{+13.90}  &
  \multicolumn{1}{c}{\textbf{+11.40}}  &
  \multicolumn{1}{c}{+3.00}&
  \multicolumn{1}{c}{\textbf{+10.70}} &
  \multicolumn{1}{c}{\textbf{+17.50}} &
   \multicolumn{1}{c}{\textbf{+10.10}} &
   \multicolumn{1}{c}{+3.30} &
   \multicolumn{1}{c}{\textbf{+3.60}} &
   \multicolumn{1}{c}{+9.30} &
   \multicolumn{1}{c}{+10.40}&
  \multicolumn{1}{c}{+2.50} &
  \multicolumn{1}{c}{+12.30} &
   \multicolumn{1}{c}{-6.10}&
   \multicolumn{1}{c}{-1.80}&
   \\
\multicolumn{1}{c|}{\multirow{-6}{*}{\begin{tabular}[c]{@{}c@{}}Zero-shot \\ with bridge\end{tabular}}} &
  \multicolumn{1}{l|}{Ours} &
   \multicolumn{1}{c}{\textbf{+10.80}}  &
  \multicolumn{1}{c}{+12.60}  &
  \multicolumn{1}{c}{\textbf{+11.40}}  &
  \multicolumn{1}{c}{+4.80}&
  \multicolumn{1}{c}{\textbf{+10.70}} &
  \multicolumn{1}{c}{\textbf{+17.50}} &
   \multicolumn{1}{c}{\textbf{+10.10}} &
   \multicolumn{1}{c}{+3.30} &
   \multicolumn{1}{c}{\textbf{+3.60}} &
   \multicolumn{1}{c}{\textbf{+16.60}} &
   \multicolumn{1}{c}{\textbf{+11.70}}&
  \multicolumn{1}{c}{\textbf{+4.10}} &
  \multicolumn{1}{c}{+14.90} &
   \multicolumn{1}{c}{\textbf{-1.30}}&
   \multicolumn{1}{c}{-2.60}& \\ 
   \bottomrule[1pt]
 \multicolumn{17}{c}{ \Large Mistral-7B} \rule{0pt}{14pt}\\ \hline
\multicolumn{2}{c|}{\textbf{Method}} &
  \multicolumn{1}{c}{{\textbf{Zh-Ja}}} &
  \multicolumn{1}{c}{{\textbf{Zh-He}}} &
  \multicolumn{1}{c}{{\textbf{Zh-Tl}}} &
  \multicolumn{1}{c}{{\textbf{Zh-Sw}}} &
  \multicolumn{1}{c}{{\textbf{Ar-Ja}}} &
  \multicolumn{1}{c}{{\textbf{Ar-He}}} &
  \multicolumn{1}{c}{{\textbf{Ar-Tl}}} &
  \multicolumn{1}{c}{{\textbf{Ar-Sw}}} &
  \multicolumn{1}{c}{{\textbf{Id-Ja}}} &
  \multicolumn{1}{c}{{\textbf{Id-He}}} &
  \multicolumn{1}{c}{{\textbf{Id-Tl}}} &
  \multicolumn{1}{c}{\textbf{Id-Sw}} &
  \multicolumn{1}{c}{\textbf{En-He}} &
  \multicolumn{1}{c}{\textbf{En-Tl}} &
  \multicolumn{1}{c}{\textbf{En-Sw}} \\ \hline
\multicolumn{2}{c|}{Zero-shot} &
  \multicolumn{1}{c}{57.80} &
  \multicolumn{1}{c}{26.20} &
  \multicolumn{1}{c}{34.10} &
  \multicolumn{1}{c}{8.40} &
  \multicolumn{1}{c}{52.50} &
  \multicolumn{1}{c}{32.30} &
  \multicolumn{1}{c}{28.00} &
  \multicolumn{1}{c}{9.10} &
  \multicolumn{1}{c}{48.40} &
  \multicolumn{1}{c}{36.20}&
  \multicolumn{1}{c}{40.60}&
   \multicolumn{1}{c}{8.60}&
\multicolumn{1}{c}{47.80}&
  \multicolumn{1}{c}{45.70}&
  \multicolumn{1}{c}{8.20} & \\ \hline
\multicolumn{1}{c|}{} &
  \multicolumn{1}{l|}{One-shot} &
  \multicolumn{1}{c}{-7.20} &
  \multicolumn{1}{c}{-0.40} &
  \multicolumn{1}{c}{-1.90} &
 \multicolumn{1}{c}{+1.60} &
  \multicolumn{1}{c}{-8.10} &
  \multicolumn{1}{c}{-7.40} &
  \multicolumn{1}{c}{-2.00} &
   \multicolumn{1}{c}{+2.10} &
   \multicolumn{1}{c}{+5.60}&
  \multicolumn{1}{c}{+2.50} &
  \multicolumn{1}{c}{-3.70} &
  \multicolumn{1}{c}{+1.20} &
  \multicolumn{1}{c}{+0.40} &
  \multicolumn{1}{c}{-13.80} &
  \multicolumn{1}{c}{+2.40} &
   \\
\multicolumn{1}{c|}{} &
  \multicolumn{1}{l|}{Two-shot} &
  \multicolumn{1}{c}{+0.40} &
  \multicolumn{1}{c}{+0.80} &
  \multicolumn{1}{c}{-2.90} &
  \multicolumn{1}{c}{+2.10} &
  \multicolumn{1}{c}{-5.10} &
   \multicolumn{1}{c}{-7.00} &
   \multicolumn{1}{c}{-0.40} &
  \multicolumn{1}{c}{+1.60}  &
    \multicolumn{1}{c}{+10.10}&
   \multicolumn{1}{c}{+3.20} &
    \multicolumn{1}{c}{-0.60}&
   \multicolumn{1}{c}{+1.60}&
   \multicolumn{1}{c}{+0.50}&
   \multicolumn{1}{c}{-3.90}&
   \multicolumn{1}{c}{\textbf{+3.20}} &
   \\
\multicolumn{1}{c|}{} &
  \multicolumn{1}{l|}{Three-shot} &
  \multicolumn{1}{c}{+3.00} &
  \multicolumn{1}{c}{+0.70} &
  \multicolumn{1}{c}{-1.60} &
  \multicolumn{1}{c}{+2.20} &
  \multicolumn{1}{c}{\textbf{-2.90}} &
  \multicolumn{1}{c}{\textbf{-6.50}} &
  \multicolumn{1}{c}{\textbf{+0.20}} &
   \multicolumn{1}{c}{+2.00}&
   \multicolumn{1}{c}{\textbf{+10.70}} &
   \multicolumn{1}{c}{+3.30}&
   \multicolumn{1}{c}{+0.80}&
   \multicolumn{1}{c}{+2.00}&
   \multicolumn{1}{c}{0.00}  &
   \multicolumn{1}{c}{-2.80}&
   \multicolumn{1}{c}{+3.00} &
   \\
\multicolumn{1}{c|}{\multirow{-4}{*}{Few-shot}} &
  \multicolumn{1}{l|}{Four-shot} &
  \multicolumn{1}{c}{\textbf{+3.20}} &
  \multicolumn{1}{c}{\textbf{+1.20}} &
  \multicolumn{1}{c}{\textbf{-0.50}} &
  \multicolumn{1}{c}{\textbf{+2.60}} &
  \multicolumn{1}{c}{\textbf{-2.90}} &
  \multicolumn{1}{c}{-6.60} &
  \multicolumn{1}{c}{0.00} &
  \multicolumn{1}{c}{\textbf{+2.20}} &
  \multicolumn{1}{c}{\textbf{+10.70}} &
  \multicolumn{1}{c}{\textbf{+4.10}} &
  \multicolumn{1}{c}{\textbf{+2.50}} &
  \multicolumn{1}{c}{\textbf{+2.60}} &
  \multicolumn{1}{c}{\textbf{+1.20}} &
  \multicolumn{1}{c}{\textbf{-2.00}} &
  \multicolumn{1}{c}{+2.50} &
   \\ \hline
\multicolumn{1}{c|}{} &
  \multicolumn{1}{l|}{Ph.D Source} &
   \multicolumn{1}{c}{-3.60}&
   \multicolumn{1}{c}{-0.20}  &
   \multicolumn{1}{c}{+0.90} &
   \multicolumn{1}{c}{+0.90}&
   \multicolumn{1}{c}{-8.70}&
   \multicolumn{1}{c}{-}&
   \multicolumn{1}{c}{-2.20}&
   \multicolumn{1}{c}{+0.80}&
   \multicolumn{1}{c}{-5.70}&
   \multicolumn{1}{c}{-5.70}&
   \multicolumn{1}{c}{-}&
   \multicolumn{1}{c}{-0.40}&
    \multicolumn{1}{c}{\textbf{-0.10}}&
    \multicolumn{1}{c}{+2.30}&
    \multicolumn{1}{c}{+2.10}&
   \\
\multicolumn{1}{c|}{} &
  \multicolumn{1}{l|}{Ph.D Target} &
   \multicolumn{1}{c}{-3.60}&
   \multicolumn{1}{c}{-1.60}&
   \multicolumn{1}{c}{-0.60}&
   \multicolumn{1}{c}{+0.20}&
   \multicolumn{1}{c}{-9.50}&
   \multicolumn{1}{c}{-}&
   \multicolumn{1}{c}{+4.90}&
   \multicolumn{1}{c}{-}&
   \multicolumn{1}{c}{-4.70}&
   \multicolumn{1}{c}{-7.20}&
   \multicolumn{1}{c}{-}&
   \multicolumn{1}{c}{-0.20}&
    \multicolumn{1}{c}{-9.00}&
    \multicolumn{1}{c}{+1.40}&
    \multicolumn{1}{c}{+1.20}&
   \\
\multicolumn{1}{c|}{} &
  \multicolumn{1}{l|}{English Bridge} &
  \multicolumn{1}{c}{\textbf{+7.90}}  &
  \multicolumn{1}{c}{\textbf{+8.60}}  &
  \multicolumn{1}{c}{\textbf{+6.40}}  &
  \multicolumn{1}{c}{+1.20} &
  \multicolumn{1}{c}{\textbf{+8.90}} &
  \multicolumn{1}{c}{\textbf{+2.70}}  &
   \multicolumn{1}{c}{\textbf{+8.60}} &
   \multicolumn{1}{c}{+1.20} &
   \multicolumn{1}{c}{\textbf{+9.60}} &
   \multicolumn{1}{c}{\textbf{+2.20}} &
   \multicolumn{1}{c}{+2.50} &
   \multicolumn{1}{c}{+0.20}&
    \multicolumn{1}{c}{-}&
  \multicolumn{1}{c}{-} &
  \multicolumn{1}{c}{-} &
   \\
\multicolumn{1}{c|}{} &
  \multicolumn{1}{l|}{Sharing Matters} &
   \multicolumn{1}{c}{\textbf{+7.90}}&
   \multicolumn{1}{c}{\textbf{+8.60}}&
   \multicolumn{1}{c}{\textbf{+6.40}}&
   \multicolumn{1}{c}{+1.20}&
   \multicolumn{1}{c}{+4.60}&
   \multicolumn{1}{c}{\textbf{+2.70}}&
   \multicolumn{1}{c}{\textbf{+8.60}}&
   \multicolumn{1}{c}{+1.20}&
   \multicolumn{1}{c}{\textbf{+9.60}}&
  \multicolumn{1}{c}{\textbf{+2.20}} &
  \multicolumn{1}{c}{+2.50} &
  \multicolumn{1}{c}{+0.20} &
  \multicolumn{1}{c}{\textbf{-0.10}} &
  \multicolumn{1}{c}{+1.50} &
  \multicolumn{1}{c}{+0.90} &
   \\
\multicolumn{1}{c|}{} &
  \multicolumn{1}{l|}{IoU Score} &
  \multicolumn{1}{c}{+0.50} &
   \multicolumn{1}{c}{+5.30}&
   \multicolumn{1}{c}{+4.90}&
   \multicolumn{1}{c}{\textbf{+2.20}}&
   \multicolumn{1}{c}{+5.90}&
   \multicolumn{1}{c}{+1.10}&
   \multicolumn{1}{c}{+7.50}&
   \multicolumn{1}{c}{\textbf{+1.60}}&
   \multicolumn{1}{c}{+2.40}&
   \multicolumn{1}{c}{-2.80}&
   \multicolumn{1}{c}{-0.50}&
   \multicolumn{1}{c}{+1.00}&
   \multicolumn{1}{c}{-1.00}&
   \multicolumn{1}{c}{+1.50}&
   \multicolumn{1}{c}{\textbf{+2.20}}&
   \\
\multicolumn{1}{c|}{} &
  \multicolumn{1}{l|}{LAPE\_{\text{overlap}}} &
   \multicolumn{1}{c}{+0.50}  &
  \multicolumn{1}{c}{+3.50}  &
  \multicolumn{1}{c}{+2.20}  &
  \multicolumn{1}{c}{\textbf{+2.20}} &
  \multicolumn{1}{c}{\textbf{+8.90}} &
  \multicolumn{1}{c}{\textbf{+2.70}}  &
   \multicolumn{1}{c}{+7.60} &
   \multicolumn{1}{c}{+1.20} &
   \multicolumn{1}{c}{\textbf{+9.60}} &
   \multicolumn{1}{c}{\textbf{+2.20}} &
   \multicolumn{1}{c}{+1.00} &
   \multicolumn{1}{c}{+0.20}&
    \multicolumn{1}{c}{\textbf{-0.10}}&
  \multicolumn{1}{c}{\textbf{+4.50}} &
  \multicolumn{1}{c}{+2.00} &
   \\
\multicolumn{1}{c|}{\multirow{-6}{*}{\begin{tabular}[c]{@{}c@{}}Zero-shot \\ with bridge\end{tabular}}} &
  \multicolumn{1}{l|}{Ours} &
  \multicolumn{1}{c}{\textbf{+7.90}}  &
  \multicolumn{1}{c}{\textbf{+8.60}}  &
  \multicolumn{1}{c}{\textbf{+6.40}}  &
  \multicolumn{1}{c}{+1.20} &
  \multicolumn{1}{c}{\textbf{+8.90}}  &
   \multicolumn{1}{c}{\textbf{+2.70}} &
   \multicolumn{1}{c}{\textbf{+8.60}} &
   \multicolumn{1}{c}{+1.20} &
   \multicolumn{1}{c}{+2.40} &
   \multicolumn{1}{c}{-3.30} &
    \multicolumn{1}{c}{\textbf{+3.30}}&
\multicolumn{1}{c}{\textbf{+1.00}}&
 \multicolumn{1}{c}{\textbf{-0.10}}&
  \multicolumn{1}{c}{+1.50}&
  \multicolumn{1}{c}{+2.00}
   &\\ 
\bottomrule[1pt]
\end{tabular}}
\end{table*}

\subsubsection{Cross-lingual Results Analysis}

Table \ref{table:BLI} compares the performance of BridgeX-ICL against
various baselines on the BLI task across $15$ language pairs.

We observe the following findings:
\textbf{1) LLMs exhibit poor and imbalanced performance on low-resource languages.}
For example,
LLaMA 3 achieves its best BLI performance of $69.90$ on 
the Ar-Ja pair, but its worst of $25.90$ 
on the Id-Sw pair.
\textbf{2) LLMs are few-shot multilingual learners},
as also noted in \cite{a19}.
However, 
few-shot X-ICL does not consistently yield stable gains.
For example, one-shot sometimes performs worse than zero-shot, 
and performance often stabilizes or may decline
once the number of shots exceeds $3$.
This suggests that when applying few-shot X-ICL, 3-shots will be enough.
\textbf{3) Zero-shot with bridge is a simple yet
data-efficient strategy for low-resource languages}.
BridgeX-ICL finds $9$ optimal bridges out of 
$15$ language pairs,
achieving average performance of two-shot X-ICL
across all pairs,
followed by the English-bridge method.
While phylogenetic distance source/target methods are the least effective.
It seems using English as the default bridge is cost-effective, which will be discussed in section \ref{dis:bridge}.

\section{Discussion}
\subsection{Candidate Bridge
Language Selection}
\label{dis:candidate}
To validate the rationale for
selecting 
Indo-European languages as 
bridge candidates,
Table \ref{tab: bridge} presents an exploratory experiment 
in which all languages in our study were
evaluated as potential bridges,
using $6$ representative language pairs. 

It shows that Indo-European languages on average  outperform the nine non-Indo-European languages. Interestingly, some non-Latin-script languages, such as Chinese, also demonstrate potential as effective bridges. The results provided valuable insights: only languages well supported by LLMs functioned as effective bridges. Based on this observation, we selected $6$ languages in Indo-European as the final candidates in our study.

\begin{table*}[!ht]
\centering
\caption{
Candidate bridge language selection comparing Indo-European languages with $9$ non-Indo-European languages.
'-'  indicates the selected bridge is either the source or target language.}
\resizebox{\textwidth}{!}{
\footnotesize 
\begin{tabular}{l|ccccccccccc}
\toprule
\textbf{} & \textbf{Zero-shot} & \textbf{Zh} & \textbf{Ja} & \textbf{Ar} & \textbf{He} & \textbf{Id} & \textbf{Tl} & \textbf{Fi} & \textbf{Hu} & \textbf{Sw} & \textbf{Indo-European (Avg.)} \\
\midrule
\textbf{Zh-Ja} & 67.10 & - & - & +3.30 & +0.80 & 0 & -5.10 & +2.80 & +5.20 & -9.90 & \textbf{+9.43} \\
\textbf{Zh-He} & 44.10 & - & +7.90 & +9.60 & - & +4.80 & -0.20 & +6.70 & +10.80 & -8.80 & \textbf{+14.37} \\
\textbf{Ar-Tl} & 46.70 & +4.50 & +3.10 & - & -7.40 & -7.10 & - & +3.40 & +4.90 & -6.10 & \textbf{+6.88} \\
\textbf{Ar-He} & 47.00 & +14.90 & +11.70 & - & - & -5.20 & +1.40 & +11.60 & +15.30 & -3.00 & \textbf{+15.72} \\
\textbf{Id-Ja} & 62.50 & -2.40 & - & -9.00 & -5.20 & - & -11.90 & -12.10 & -9.60 & -24.80 & \textbf{+1.40} \\
\textbf{Id-Sw} & 25.90 & +3.20 & +1.50 & +1.20 & -0.80 & - & -3.90 & -0.50 & +1.30 & - & \textbf{+3.43} \\
\bottomrule
\end{tabular}}
\label{tab: bridge}
\end{table*}

\subsection{Application of Bridge Language}
\label{dis:bridge} 
Beyond the BLI task,
Table \ref{bridgeapp} evaluates BridgeX-ICL on the MRC cross-lingual task. 
The prompt for zero-shot with bridge in MRC is detailed in Appendix~\ref{sec:appendixC_prompt}. 
Appendix~\ref{sec: task} further evaluates BridgeX-ICL on CLQA and XNLI cross-lingual tasks. 
Results show our approach works well and benefits more from LLaMA 3 than from Mistral.
For example, BridgeX-ICL improves the performance of LLaMA 3 by an average of $6.03$\%  over the zero-shot baseline across $15$ language pairs, 
while the average improvement on Mistral is $4.48$\%.

English is selected as
the optimal bridge in $9$ out of $15$ language pairs in LLaMA 3 ($6$ out of $15$ in Mistral).
This is partly due to the unbalanced language abilities of LLMs across 5 candidate  
Indo-European bridges.
As discussed in Figure \ref{fig:pathvisuall}, 
another key factor is the model's inherent
preference for English-pivot during cross-lingual transfer.

\begin{table}[!ht]
\caption{Evaluation on MRC cross-lingual task. 
\textcolor{red}{Red} color highlights the different bridge selections, and \textbf{bold} marks the highest gains at each language pair.}
\label{bridgeapp}
\setlength{\arrayrulewidth}{0.5pt} 
\renewcommand{\arraystretch}{1.1} 
\resizebox{0.48\textwidth}{!}{
\footnotesize
\begin{tabular}{l|ccc|ccc} 
\toprule[0.5pt]
& \multicolumn{3}{c|}{LLaMA-3-8B} & \multicolumn{3}{c}{Mistral-7B} \\
\cline{2-7}
 & Bridge & Zero-shot & Ours & Bridge & Zero-shot & Ours \\
\hline
\textbf{Zh-Ja} & En & 61.80 & -0.40 & En  & 66.20 & \textbf{+6.20} \\
\textbf{Zh-He} & En & 56.00 & +7.20 & En & 50.20 & \textbf{+10.20} \\
\textbf{Zh-Tl} & En & 57.20 & +3.00 & En & 60.60 & \textbf{+7.80} \\
\textbf{Zh-Sw} & \textcolor{red}{En} & 48.60 & \textbf{+10.40} & \textcolor{red}{Pt} & 43.00 & +1.00 \\
\textbf{Ar-Ja} & En & 52.20 & +3.20 & En & 48.40 & \textbf{+7.40} \\
\textbf{Ar-He} & En & 51.20 & +4.00 & En & 42.00 & \textbf{+5.20} \\
\textbf{Ar-Tl} & En & 46.40 & \textbf{+7.20} & En & 47.60 & +4.60 \\
\textbf{Ar-Sw} & En & 40.60 & \textbf{+12.60} & En & 30.40 & +9.80 \\
\textbf{Id-Ja} & \textcolor{red}{En} & 56.20 & \textbf{+8.20} & \textcolor{red}{Pt} & 63.40 & +2.00 \\
\textbf{Id-He} & Es & 58.20 & \textbf{+7.00} & Es & 45.80 & +6.40 \\
\textbf{Id-Tl} & \textcolor{red}{Fr} & 58.40 & \textbf{+4.20} & \textcolor{red}{De} & 55.60 & +3.00 \\
\textbf{Id-Sw} & \textcolor{red}{Fr} & 49.80 & \textbf{+5.20} & \textcolor{red}{Pt} & 39.60 & -2.80 \\
\textbf{En-He} & Es & 70.60 & \textbf{+8.00} & Es & 61.20 & +3.80 \\
\textbf{En-Tl} & \textcolor{red}{Fr} & 72.80 & \textbf{+3.20} & \textcolor{red}{Es} & 72.20 & +2.40 \\
\textbf{En-Sw} & Fr & 64.60 & \textbf{+7.40} & Fr & 51.20 & +0.20 \\
\bottomrule[0.5pt]
\end{tabular}
}
\end{table}

\subsection{Ablation Study}
In this part, we conduct ablation study to evaluate the impact of the constructed neuron probe data and the proposed HSIC similarity metric on bridge selection, using the BLI task as an example.

\begin{table}[!h]
\centering
\caption{The impact of neuron probe data. `w/o $*$' denotes replacing our constructed probe data with bilingual tokens extracted from the FLORES+ dataset.} 
\label{ablation-pro}
\setlength{\arrayrulewidth}{0.5pt}
\renewcommand{\arraystretch}{1.1}
\resizebox{0.4\textwidth}{!}{ 
\scriptsize 
\begin{tabular}{l|lr|lr} 
\toprule[0.5pt]
 & \multicolumn{2}{c|}{LLaMA-3-8B} & \multicolumn{2}{c}{Mistral-7B} \\
\cline{2-5}
 & \multicolumn{1}{l}{~w/o $*$~} & \multicolumn{1}{l|}{~~Ours~~} & \multicolumn{1}{l}{~w/o $*$~} & \multicolumn{1}{l}{~~Ours~~} \\
\hline
\textbf{Zh-Ja} & ~76.40~ & ~77.90~\,\textcolor{green}{$\uparrow$}~ & ~63.40~ & ~65.70~\,\textcolor{green}{$\uparrow$}~ \\
\textbf{Zh-He} & ~56.70~ & ~56.70~\,\textcolor{blue}{--}~ & ~32.70~ & ~34.80~\,\textcolor{green}{$\uparrow$}~ \\
\textbf{Zh-Tl} & ~51.10~ & ~54.00~\,\textcolor{green}{$\uparrow$}~ & ~38.20~ & ~40.50~\,\textcolor{green}{$\uparrow$}~ \\
\textbf{Zh-Sw} & ~36.40~ & ~36.00~\,\textcolor{red}{$\downarrow$}~  & ~10.60~ & ~9.60~\,\textcolor{red}{$\downarrow$}~ \\
\textbf{Ar-Ja} & ~77.80~ & ~80.60~\,\textcolor{green}{$\uparrow$}~ & ~58.40~ & ~61.40~\,\textcolor{green}{$\uparrow$}~ \\
\textbf{Ar-He} & ~64.10~ & ~64.50~\,\textcolor{green}{$\uparrow$}~ & ~33.40~ & ~35.00~\,\textcolor{green}{$\uparrow$}~ \\
\textbf{Ar-Tl} & ~52.00~ & ~56.80~\,\textcolor{green}{$\uparrow$}~ & ~33.30~ & ~35.50~\,\textcolor{green}{$\uparrow$}~ \\
\textbf{Ar-Sw} & ~41.60~ & ~42.40~\,\textcolor{green}{$\uparrow$}~ & ~10.70~ & ~10.70~\,\textcolor{blue}{--}~ \\
\textbf{Id-Ja} & ~65.10~ & ~66.10~\,\textcolor{green}{$\uparrow$}~ & ~50.80~ & ~58.00~\,\textcolor{green}{$\uparrow$}~ \\
\textbf{Id-He} & ~59.90~ & ~61.30~\,\textcolor{green}{$\uparrow$}~ & ~33.40~ & ~38.40~\,\textcolor{green}{$\uparrow$}~ \\
\textbf{Id-Tl} & ~57.00~ & ~61.00~\,\textcolor{green}{$\uparrow$}~ & ~40.20~ & ~40.20~\,\textcolor{blue}{--}~ \\
\textbf{Id-Sw} & ~28.30~ & ~30.00~\,\textcolor{green}{$\uparrow$}~ & ~~~9.20~ & ~9.20~\,\textcolor{blue}{--}~ \\
\textbf{En-He} & ~66.60~ & ~71.80~\,\textcolor{green}{$\uparrow$}~ & ~47.70~ & ~47.70~\,\textcolor{blue}{--}~ \\
\textbf{En-Tl} & ~58.70~ & ~58.70~\,\textcolor{blue}{--}~ & ~47.70~ & ~48.00~\,\textcolor{green}{$\uparrow$}~ \\
\textbf{En-Sw} & ~26.20~ & ~26.20~\,\textcolor{blue}{--}~ & ~10.20~ & ~10.30~\,\textcolor{green}{$\uparrow$}~ \\
\bottomrule[0.5pt]
\end{tabular}
}
\end{table}

Table \ref{ablation-pro} 
presents the ablation results by comparing 
``w/o $*$'' with our constructed 
probe data,
where ``w/o $*$'' denotes replacing 
the constructed probe data 
with the bilingual tokens 
extracted from FLORES+ \cite{x7}.
The detailed construction of ``w/o $*$'' is 
provided in 
Appendix~\ref{sec:appendixBAlare}.
The results highlight the crucial impact of probe data on
effective neuron manipulation.
Furthermore, Table \ref{hsicaba} in Appendix~\ref{sec:appendixBAlare} compares 
HSIC with Cosine similarity,
showing that HSIC better captures the dependency 
between language-overlapping neurons and specific neurons.

\subsection{Overlapping Neuron Distribution}

\begin{figure}[!h]
\centering
\includegraphics[width=0.48\textwidth]
{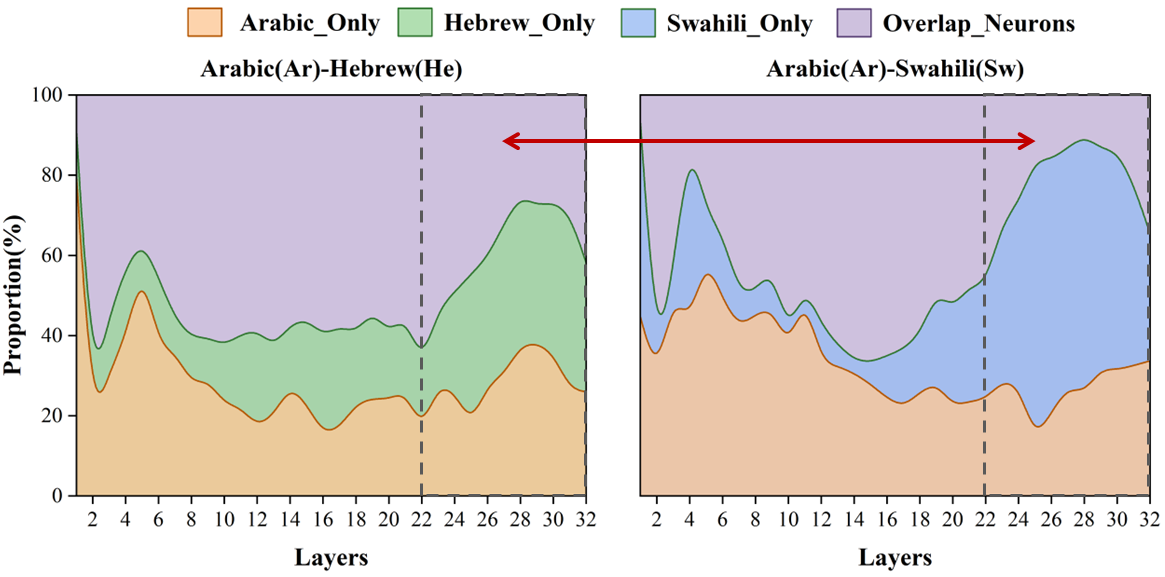}
\caption{
\textbf{Distribution of overlap neurons} in language pairs
within and across families
in LLaMA 3.}
\label{fig:no_overlapl}
\end{figure}

\begin{figure}[!h]
\centering
\includegraphics[width=0.5\textwidth]{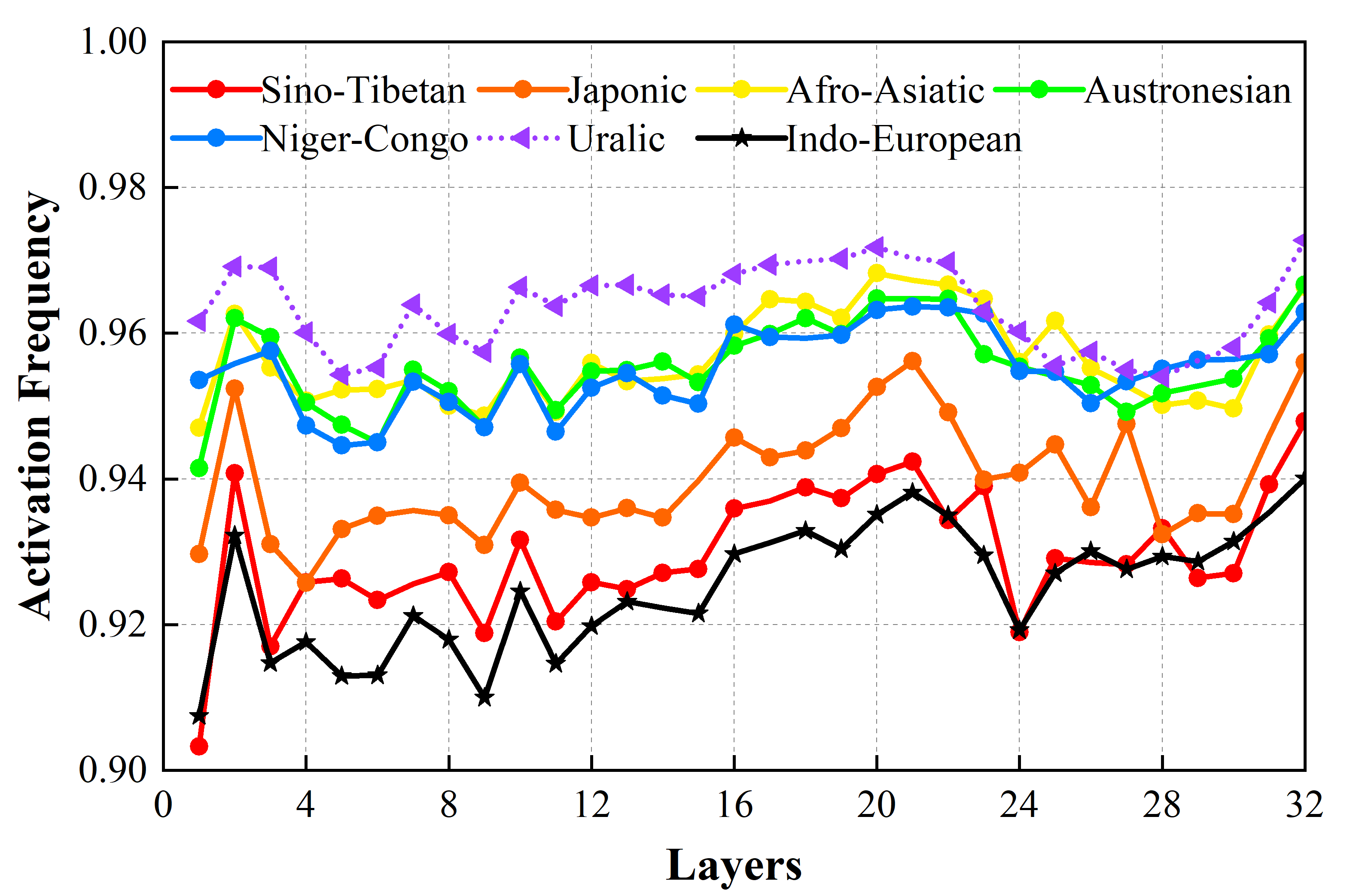}     
\caption{
\textbf{Layer-wise activation frequency of neurons}
viewed from language families in LLaMA 3.}
\label{fig:activefam}
\end{figure}

This section analyzes the distribution of 
overlap neurons.
Figure \ref{fig:no_overlapl} 
compares neurons in language pairs within the same family (e.g., Ar-He) 
and across families (e.g., Ar-Sw).
Obviously, 
Ar-He shares more overlapping neurons.
Similar observations can be found 
in comparing language pairs from 
different source languages 
to a same target language (Figure \ref{fig:no_fresss} in Appendix~\ref{sec:appendixond})

From the perspective of language families,
Figure \ref{fig:activefam} examines the activated behaviors of neurons in low-resource languages.
Obviously,
low-resource languages within the
Uralic family have the highest activation frequency,
while Indo-European languages have the lowest.
We hypothesize LLMs activate neurons more frequently for processing low-resource languages due to their perceived difficulty.

\section{Conclusion}
In this work,
we explore whether sharing neurons can improve LLMs' cross-lingual performance on low-resource languages.
We propose a simple yet effective language-bridge approach
with the help of neuron interpretation.
To ensure accurate and full activation of
overlap neurons across languages,
we construct probing data
from the ground-truth MUSE dictionaries.
By quantifying neuron similarity,
we seek the optimal bridge for X-ICL and
conduct extensive experiments 
to validate its efficacy and generalization.

\section*{Limitations}
This work focuses on sharing neurons across languages
and relies on the evaluated
datasets to validate the effectiveness of our approach.
Due to the lack of comprehensive benchmarks for low-resource languages,
our experiments cover only
$15$ language pairs
and select $4$ low-resource languages from distinct families as target languages and test their performance on $4$ cross-lingual tasks. 
Second, our study reveals that high-quality probe data is crucial to accurately analyze neuron behaviors of low-resource languages,
while  
the proposed linguistic distance 
measurement is probe-data-induced,
offering qualitative but not quantitative insights. 
Finally, although bridge selection should ideally follow linguistic phylogeny, we aim to select bridges that LLMs can best exploit,  inevitably reflecting their training biases.

\bibliography{main}

\begin{thebibliography}{38}
\providecommand{\natexlab}[1]{#1}

\bibitem[{Al~Nazi et~al.(2024)Al~Nazi, Hossain, and Al~Mamun}]{a19}
Zabir Al~Nazi, Md~Rajib Hossain, and Faisal Al~Mamun. 2024.
\newblock Evaluation of open and closed-source llms for a low-resource language with zero-shot, few-shot , and chain-of-thought prompting.

\bibitem[{Bandarkar et~al.(2024)Bandarkar, Liang, Muller, Artetxe, Shukla, Husa, Goyal, Krishnan, Zettlemoyer, and Khabsa}]{x11}
Lucas Bandarkar, Davis Liang, Benjamin Muller, Mikel Artetxe, Satya~Narayan Shukla, Donald Husa, Naman Goyal, Abhinandan Krishnan, Luke Zettlemoyer, and Madian Khabsa. 2024.
\newblock The belebele benchmark: a parallel reading comprehension dataset in 122 language variants.
\newblock In \emph{Proceedings of the 62nd Annual Meeting of the Association for Computational Linguistics (Volume 1: Long Papers)}, pages 749--775, Bangkok, Thailand and virtual meeting. Association for Computational Linguistics.

\bibitem[{Brown et~al.(2020{\natexlab{a}})Brown, Mann, Ryder, Subbiah, Kaplan, Dhariwal, Neelakantan, Shyam, Sastry, Askell et~al.}]{a32}
Tom Brown, Benjamin Mann, Nick Ryder, Melanie Subbiah, Jared~D Kaplan, Prafulla Dhariwal, Arvind Neelakantan, Pranav Shyam, Girish Sastry, Amanda Askell, and 1 others. 2020{\natexlab{a}}.
\newblock Language models are few-shot learners.
\newblock \emph{Advances in neural information processing systems}, 33:1877--1901.

\bibitem[{Brown et~al.(2020{\natexlab{b}})Brown, Mann, Ryder, Subbiah, Kaplan et~al.}]{a11}
Tom~B. Brown, Benjamin Mann, Nick Ryder, Melanie Subbiah, Jared Kaplan, and 1 others. 2020{\natexlab{b}}.
\newblock Language models are probing learners.
\newblock In \emph{NeurIPS}.

\bibitem[{Cahyawijaya et~al.(2024)Cahyawijaya, Lovenia, and Fung}]{a29}
Samuel Cahyawijaya, Holy Lovenia, and Pascale Fung. 2024.
\newblock Llms are few-shot in-context low-resource language learners.
\newblock In \emph{Proceedings of the 2024 Conference of the North American Chapter of the Association for Computational Linguistics: Human Language Technologies (Volume 1: Long Papers)}, pages 405--433.

\bibitem[{Cao et~al.(2024)Cao, Chen, Jin, Chen, Liu, and Zhao}]{a31}
Pengfei Cao, Yuheng Chen, Zhuoran Jin, Yubo Chen, Kang Liu, and Jun Zhao. 2024.
\newblock One mind, many tongues: A deep dive into language-agnostic knowledge neurons in large language models.
\newblock \emph{arXiv preprint arXiv:2411.17401}.

\bibitem[{Cieri et~al.(2016)Cieri, Maxwell, Strassel, and Tracey}]{a34}
Christopher Cieri, Mike Maxwell, Stephanie Strassel, and Jennifer Tracey. 2016.
\newblock Selection criteria for low resource language programs.
\newblock In \emph{Proceedings of the Tenth International Conference on Language Resources and Evaluation (LREC'16)}, pages 4543--4549.

\bibitem[{Conneau et~al.(2020)Conneau, Khandelwal, Goyal, Chaudhary, Wenzek, Guzm{\'a}n, Grave, Ott, Zettlemoyer, and Stoyanov}]{a4}
Alexis Conneau, Kartikay Khandelwal, Naman Goyal, Vishrav Chaudhary, Guillaume Wenzek, Francisco Guzm{\'a}n, {\'E}douard Grave, Myle Ott, Luke Zettlemoyer, and Veselin Stoyanov. 2020.
\newblock Unsupervised cross-lingual representation learning at scale.
\newblock In \emph{Proceedings of the 58th Annual Meeting of the Association for Computational Linguistics}, pages 8440--8451.

\bibitem[{Conneau et~al.(2017)Conneau, Lample, Ranzato, Denoyer, and J{\'e}gou}]{x1}
Alexis Conneau, Guillaume Lample, Marc'Aurelio Ranzato, Ludovic Denoyer, and Herv{\'e} J{\'e}gou. 2017.
\newblock Word translation without parallel data.
\newblock \emph{arXiv preprint arXiv:1710.04087}.

\bibitem[{Costa-juss{\`a} et~al.(2022)Costa-juss{\`a}, Cross, {\c{C}}elebi, Elbayad, Heafield, Heffernan, Kalbassi, Lam, Licht, Maillard et~al.}]{a35}
Marta~R Costa-juss{\`a}, James Cross, Onur {\c{C}}elebi, Maha Elbayad, Kenneth Heafield, Kevin Heffernan, Elahe Kalbassi, Janice Lam, Daniel Licht, Jean Maillard, and 1 others. 2022.
\newblock No language left behind: Scaling human-centered machine translation.
\newblock \emph{arXiv preprint arXiv:2207.04672}.

\bibitem[{Dubey et~al.(2024)Dubey, Jauhri, Pandey, Kadian, Al-Dahle, Letman, Mathur, Schelten, Yang, Fan et~al.}]{a3}
Abhimanyu Dubey, Abhinav Jauhri, Abhinav Pandey, Abhishek Kadian, Ahmad Al-Dahle, Aiesha Letman, Akhil Mathur, Alan Schelten, Amy Yang, Angela Fan, and 1 others. 2024.
\newblock The llama 3 herd of models.
\newblock \emph{arXiv preprint arXiv:2407.21783}.

\bibitem[{Foundation(2024)}]{a58}
Wikimedia Foundation. 2024.
\newblock \href {https://dumps.wikimedia.org/} {Wikimedia downloads.}

\bibitem[{Gray et~al.(2009)Gray, Drummond, and Greenhill}]{a51}
Russell~D. Gray, Alexei~J. Drummond, and Simon~J. Greenhill. 2009.
\newblock Language phylogenies reveal expansion pulses and pauses in pacific settlement.
\newblock \emph{Science}, 323(5913):479--483.

\bibitem[{Gretton et~al.(2005)Gretton, Bousquet, Smola, and Sch{\"o}lkopf}]{a59}
Arthur Gretton, Olivier Bousquet, Alex Smola, and Bernhard Sch{\"o}lkopf. 2005.
\newblock Measuring statistical dependence with hilbert-schmidt norms.
\newblock In \emph{International conference on algorithmic learning theory}, pages 63--77. Springer.

\bibitem[{Hammarström et~al.(2023)Hammarström, Forkel, Haspelmath, and Bank}]{a52}
Harald Hammarström, Robert Forkel, Martin Haspelmath, and Sebastian Bank. 2023.
\newblock Glottolog 5.1.

\bibitem[{Huang et~al.(2024)Huang, Mo, Li, Li, Zhang, Yi, Mao, Liu, Xu, Xu et~al.}]{a22}
Kaiyu Huang, Fengran Mo, Hongliang Li, You Li, Yuanchi Zhang, Weijian Yi, Yulong Mao, Jinchen Liu, Yuzhuang Xu, Jinan Xu, and 1 others. 2024.
\newblock A survey on large language models with multilingualism: Recent advances and new frontiers.
\newblock \emph{arXiv preprint arXiv:2405.10936}.

\bibitem[{Huerta-Cepas et~al.(2016)Huerta-Cepas, Serra, and Bork}]{a53}
Jaime Huerta-Cepas, François Serra, and Peer Bork. 2016.
\newblock Ete 3: Reconstruction, analysis, and visualization of phylogenomic data.
\newblock \emph{Molecular Biology and Evolution}, 33(6):1635--1638.

\bibitem[{Izbicki(2022)}]{x2}
Mike Izbicki. 2022.
\newblock Aligning word vectors on low-resource languages with {W}iktionary.
\newblock In \emph{Proceedings of the Fifth Workshop on Technologies for Machine Translation of Low-Resource Languages (LoResMT 2022)}, pages 107--117, Gyeongju, Republic of Korea. Association for Computational Linguistics.

\bibitem[{Jiang et~al.(2023)Jiang, Sablayrolles, Mensch, Bamford, Chaplot, Casas, Bressand, Lengyel, Lample, Saulnier et~al.}]{a45}
Albert~Q Jiang, Alexandre Sablayrolles, Arthur Mensch, Chris Bamford, Devendra~Singh Chaplot, Diego de~las Casas, Florian Bressand, Gianna Lengyel, Guillaume Lample, Lucile Saulnier, and 1 others. 2023.
\newblock Mistral 7b.
\newblock \emph{arXiv preprint arXiv:2310.06825}.

\bibitem[{Liu et~al.(2024)Liu, Xu, Xu, and other}]{a16}
Weize Liu, Yinlong Xu, Hongxia Xu, and other. 2024.
\newblock Unraveling babel: Exploring multilingual activation patterns of llms and their applications.
\newblock In \emph{{EMNLP}}, pages 11855--11881. Association for Computational Linguistics.

\bibitem[{McInnes et~al.(2018)McInnes, Healy, and Melville}]{x13}
Leland McInnes, John Healy, and James Melville. 2018.
\newblock Umap: Uniform manifold approximation and projection for dimension reduction.
\newblock \emph{arXiv preprint arXiv:1802.03426}.

\bibitem[{Mondal et~al.(2025)Mondal, Sen, Singhania, and Jyothi}]{a50}
Soumen~Kumar Mondal, Sayambhu Sen, Abhishek Singhania, and Preethi Jyothi. 2025.
\newblock Language-specific neurons do not facilitate cross-lingual transfer.
\newblock \emph{arXiv preprint arXiv:2503.17456}.

\bibitem[{Muennighoff et~al.(2023)Muennighoff, Rush, Barak, Le~Scao, Tazi, Piktus, Pyysalo, Wolf, and Raffel}]{a38}
Niklas Muennighoff, Alexander Rush, Boaz Barak, Teven Le~Scao, Nouamane Tazi, Aleksandra Piktus, Sampo Pyysalo, Thomas Wolf, and Colin~A Raffel. 2023.
\newblock Scaling data-constrained language models.
\newblock \emph{Advances in Neural Information Processing Systems}, 36:50358--50376.

\bibitem[{Muller et~al.(2021)Muller, Anastasopoulos, Sagot, and Seddah}]{a23}
Benjamin Muller, Antonios Anastasopoulos, Beno{\^\i}t Sagot, and Djam{\'e} Seddah. 2021.
\newblock When being unseen from mbert is just the beginning: Handling new languages with multilingual language models.
\newblock In \emph{Proceedings of the 2021 Conference of the North American Chapter of the Association for Computational Linguistics: Human Language Technologies}, pages 448--462.

\bibitem[{{NLLB Team} et~al.(2024){NLLB Team}, Costa-juss{\`a}, Cross, {\c{C}}elebi, and et~al.}]{x7}
{NLLB Team}, Marta~R. Costa-juss{\`a}, James Cross, Onur {\c{C}}elebi, and et~al. 2024.
\newblock Scaling neural machine translation to 200 languages.
\newblock \emph{Nature}, 630(8018):841--846.

\bibitem[{Nostalgebraist(2020)}]{a60}
Nostalgebraist. 2020.
\newblock Interpreting gpt: The logit lens.
\newblock LessWrong.

\bibitem[{Philippy et~al.(2023)Philippy, Guo, and Haddadan}]{a13}
Fred Philippy, Siwen Guo, and Shohreh Haddadan. 2023.
\newblock Towards a common understanding of contributing factors for cross-lingual transfer in multilingual language models: {A} review.
\newblock In \emph{Proceedings of the 61st Annual Meeting of the Association for Computational Linguistics}, pages 5877--5891.

\bibitem[{Song et~al.(2024)Song, He, Jiang, Xian, Gao, Liu, and Yu}]{a57}
Ran Song, Shizhu He, Shuting Jiang, Yantuan Xian, Shengxiang Gao, Kang Liu, and Zhengtao Yu. 2024.
\newblock Does large language model contain task-specific neurons?
\newblock In \emph{Proceedings of the 2024 Conference on Empirical Methods in Natural Language Processing}, pages 7101--7113.

\bibitem[{Stanczak et~al.(2022)Stanczak, Ponti, Hennigen, Cotterell, and Augenstein}]{a40}
Karolina Stanczak, Edoardo Ponti, Lucas~Torroba Hennigen, Ryan Cotterell, and Isabelle Augenstein. 2022.
\newblock Same neurons, different languages: Probing morphosyntax in multilingual pre-trained models.
\newblock In \emph{Proceedings of the 2022 Conference of the North American Chapter of the Association for Computational Linguistics: Human Language Technologies}, pages 1589--1598.

\bibitem[{Tan et~al.(2024)Tan, Wu, and Monz}]{x12}
Shaomu Tan, Di~Wu, and Christof Monz. 2024.
\newblock Neuron specialization: Leveraging intrinsic task modularity for multilingual machine translation.
\newblock In \emph{Proceedings of the 2024 Conference on Empirical Methods in Natural Language Processing}, pages 6506--6527, Miami, Florida, USA. Association for Computational Linguistics.

\bibitem[{Tang et~al.(2024)Tang, Luo, Huang et~al.}]{a15}
Tianyi Tang, Wenyang Luo, Haoyang Huang, and 1 others. 2024.
\newblock Language-specific neurons: The key to multilingual capabilities in large language models.
\newblock In \emph{{ACL} {(1)}}, pages 5701--5715. Association for Computational Linguistics.

\bibitem[{Tanwar et~al.(2023)Tanwar, Dutta, Borthakur, and Chakraborty}]{a28}
Eshaan Tanwar, Subhabrata Dutta, Manish Borthakur, and Tanmoy Chakraborty. 2023.
\newblock Multilingual llms are better cross-lingual in-context learners with alignment.
\newblock In \emph{Proceedings of the 61st Annual Meeting of the Association for Computational Linguistics (Volume 1: Long Papers)}, pages 6292--6307.

\bibitem[{Vulic et~al.(2020)Vulic, Ponti, Litschko, Glavas, and Korhonen}]{a21}
Ivan Vulic, Edoardo~Maria Ponti, Robert Litschko, Goran Glavas, and Anna Korhonen. 2020.
\newblock Probing pretrained language models for lexical semantics.
\newblock In \emph{{EMNLP} {(1)}}, pages 7222--7240. Association for Computational Linguistics.

\bibitem[{Wang et~al.(2024)Wang, Haddow, Wu, Peng, and Birch}]{a30}
Weixuan Wang, Barry Haddow, Minghao Wu, Wei Peng, and Alexandra Birch. 2024.
\newblock Sharing matters: Analysing neurons across languages and tasks in llms.
\newblock \emph{arXiv preprint arXiv:2406.09265}.

\bibitem[{Wichmann and Holman(2009)}]{a54}
Søren Wichmann and Eric~W. Holman. 2009.
\newblock \emph{Temporal stability of linguistic typological features}.
\newblock Lincom Europa.

\bibitem[{Winata et~al.(2021)Winata, Madotto, Lin, Liu, Yosinski, and Fung}]{a27}
Genta~Indra Winata, Andrea Madotto, Zhaojiang Lin, Rosanne Liu, Jason Yosinski, and Pascale Fung. 2021.
\newblock Language models are few-shot multilingual learners.
\newblock In \emph{Proceedings of the 1st Workshop on Multilingual Representation Learning}, pages 1--15.

\bibitem[{Yong et~al.(2023)Yong, Schoelkopf, Muennighoff, Aji, Adelani, Almubarak, Bari, Sutawika, Kasai, Baruwa et~al.}]{a26}
Zheng~Xin Yong, Hailey Schoelkopf, Niklas Muennighoff, Alham~Fikri Aji, David~Ifeoluwa Adelani, Khalid Almubarak, M~Saiful Bari, Lintang Sutawika, Jungo Kasai, Ahmed Baruwa, and 1 others. 2023.
\newblock Bloom+ 1: Adding language support to bloom for zero-shot prompting.
\newblock In \emph{Proceedings of the 61st Annual Meeting of the Association for Computational Linguistics (Volume 1: Long Papers)}, pages 11682--11703.

\bibitem[{Zhang et~al.(2025)Zhang, Chen, Bai, Li, Xiang, and Zhang}]{a61}
Hongbin Zhang, Kehai Chen, Xuefeng Bai, Xiucheng Li, Yang Xiang, and Min Zhang. 2025.
\newblock Exploring translation mechanism of large language models.
\newblock \emph{arXiv preprint arXiv:2502.11806}.

\end{thebibliography}

\newpage
\appendix
\section{Appendix: Linguistic Similarity Based on Glottolog Phylogenetic Trees }
\label{sec:appendixA}
We leverage Glottolog version 5.1 \cite{a52}  as a foundational phylogenetic framework to calculate the linguistic similarity of human languages. It has two key steps: data preprocessing and similarity calculation.

\noindent\textbf{Data Preprocessing.}
The preprocessing pipeline consists of three key steps: 1) 
Locate Glottocode identifiers with regex pattern matching to ensure unambiguous language node identification. 2) Standardize node naming with underscores 
(e.g.,  [sini1245] $\rightarrow$ \_slaini1245\_), ensuring consistent formatting in downstream phylogenetic analyses. 
3) Mitigate encoding conflicts through temporary file caching. These steps preserve accurate 
parsing of phylogenetic tree 
while maintaining computational compatibility.

\noindent\textbf{Similarity Calculation.}
The proposed metric integrates two well-established principles from historical linguistics:
node distance normalization and depth-adjusted compensation.

First,
building upon Wichmann \& Holman's framework for typological stability assessment \cite{a54},
we compute the 
inter-language distance $d(L_1,L_2)$ between languages $L_1$ and $L_2$  using ETE3's optimized tree traversal algorithms \cite{a53}.
We then normalize the distance 
to make it comparable across language families, calculated as:
\begin{equation}
S_\mathrm{distance}=1-\min\left(1,\frac{d(L_1,L_2)}{\hat{D}}\right)
\end{equation}
where $\hat{D}$ is
the family-specific  maximum.
For example, $\hat{D} = 80$ for Sino-Tibetan languages, 
reflecting their deep internal divergence,
whereas $\hat{D} = 75$
for Indo-European languages, due to their relatively shallower subgroup structure.

Second, depth-adjusted compensation
aims to mitigate biases introduced by uneven tree depth and family-specific structural variation. 
Following the work \cite{a51} to calculate depth disparity factor $\delta(L_1,L_2)$,
we measure the depth $\alpha_{\text{depth}}(L_1,L_2)$ between $L_1$ and $L_2$ as:
\begin{equation}
\alpha_{\mathrm{depth}}=1-\frac{\delta(L_1,L_2)}{\max(\text{depth}(L_1),\text{depth}(L_2))}
\end{equation}
The final language similarity score is computed as:
\begin{equation}
\mathrm{Sim}(L_1,L_2)=S_\mathrm{distance}\times\alpha_\mathrm{depth}
\end{equation}

\section{Appendix: Supplementary Cross-lingual Tasks}
\label{sec: task}
To further verify robustness, we conduct additional experiments 
using the bridge languages selected by our method on downstream tasks, including Cross-Lingual Question Answering (CLQA) and Cross-Lingual Natural Language Inference (XNLI), as shown in Table~\ref{tab:clqa}. 
Due to the limited availability of cross-lingual benchmarks covering our target low-resource languages (e.g., Tagalog), 
the evaluation is restricted to $6$ language pairs for CLQA and $3$ pairs for XNLI.

\begin{table}[!ht]
\caption{Evaluation on CLQA and XNLI cross-lingual tasks. \textbf{Bold} highlights improved performance.}
\label{tab:clqa}
\setlength{\arrayrulewidth}{0.5pt}
\renewcommand{\arraystretch}{1.1}
\resizebox{0.48\textwidth}{!}{
\footnotesize
\begin{tabular}{l|cc|cc} 
\toprule[0.5pt]
& \multicolumn{2}{c|}{CLQA} & \multicolumn{2}{c}{XNLI} \\
\cline{2-5}
 & Zero-shot & Ours & Zero-shot & Ours \\
\hline
\textbf{Zh-Ja} & 42.80 & -0.60 & - & - \\
\textbf{Zh-Sw} & 38.20 & \textbf{+5.20} & 35.90 & \textbf{+1.30} \\
\textbf{Ar-Ja} & 41.60 & \textbf{+6.20} & - & - \\
\textbf{Ar-Sw} & 34.20 & \textbf{+2.40} & 35.50 & \textbf{+3.70} \\
\textbf{Id-Ja} & 34.60 & \textbf{+3.60} & - & - \\
\textbf{Id-Sw} & 34.80 & \textbf{+5.20} & 36.60 & \textbf{+0.60} \\
\bottomrule[0.5pt]
\end{tabular}
}
\end{table}

\section{Appendix: Neuron Patterns}
\label{sec:appendixB}

\subsection{Deactivation Overlap Neurons}
\label{actiionpp}

\begin{figure}[!h]
\centering
\includegraphics[width=0.5\textwidth]{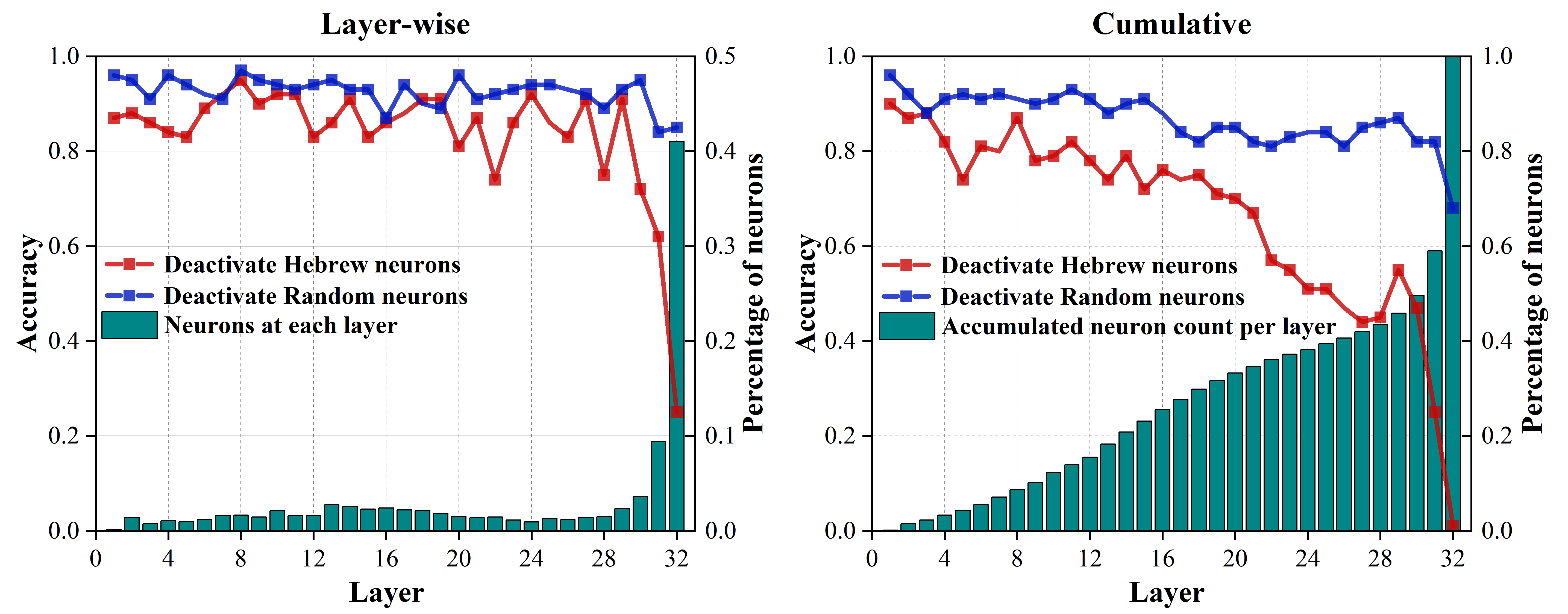}     
\caption{
\textbf{Overlap neuron distributions and their deactivation effects} on the Chinese-Hebrew BLI task.
}
\label{fig8:mistddd}
\end{figure}

Figure \ref{fig8:mistddd} presents
the distribution of overlap neurons and their deactivation effects on the Chinese-Hebrew (Zh-He) BLI task.

\subsection{Linguistic Spectrum in Mistral}
\label{sec:appendixB_Mistral}

\begin{figure}[!h]
\centering
\includegraphics[width=0.5\textwidth]{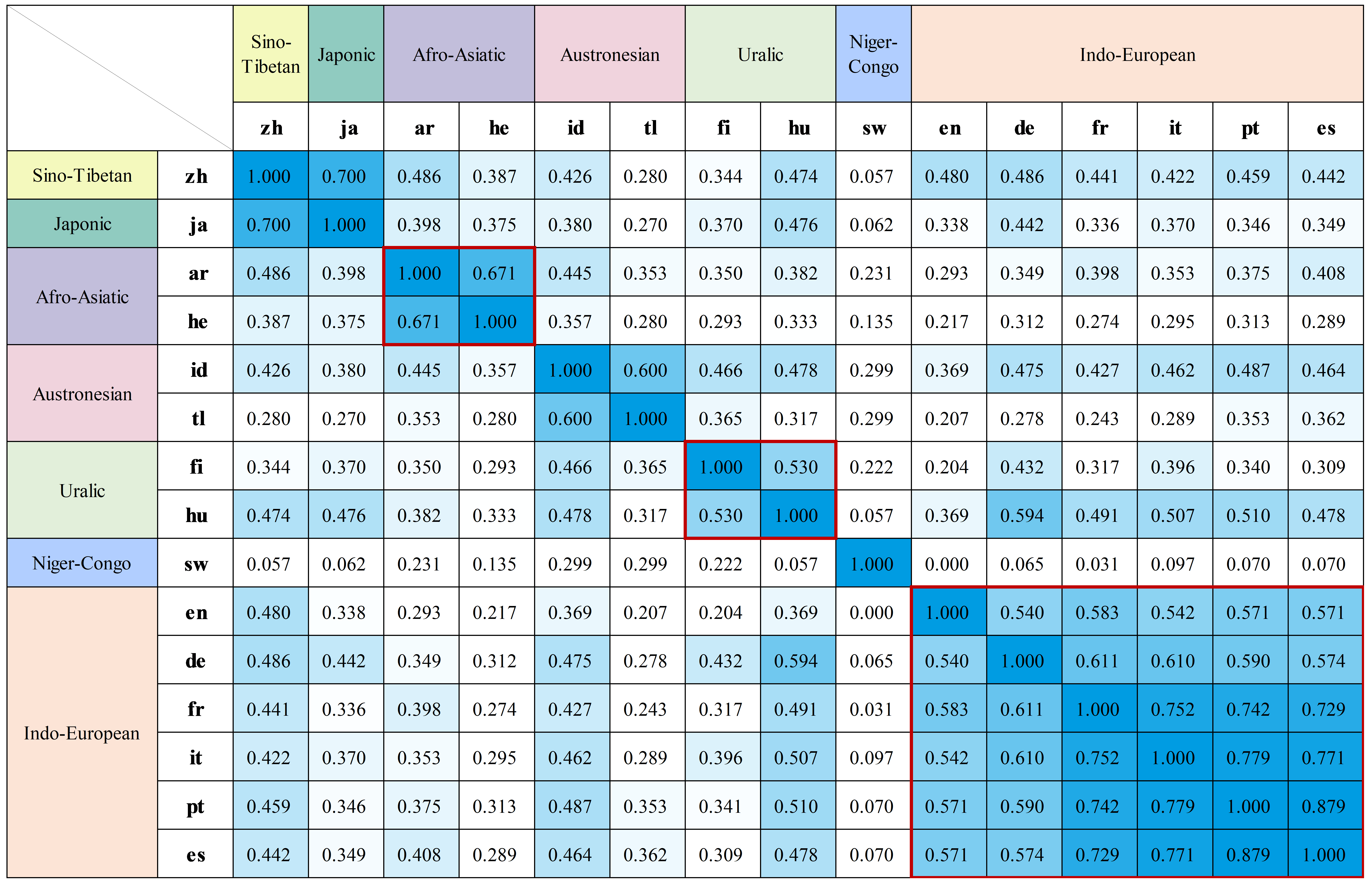}    
\caption{\textbf{Mistral's linguistic spectrum}
across $15$ languages from $7$ families. 
The color intensity represents the degree of overlap between language pairs.}
\label{fig:activemistralspr}
\end{figure}

\subsection{Parameter $K$ Discussion}
\label{sec:appendixB_k_dis}

\begin{figure}[!h]
\centering
\includegraphics[width=0.47\textwidth]
{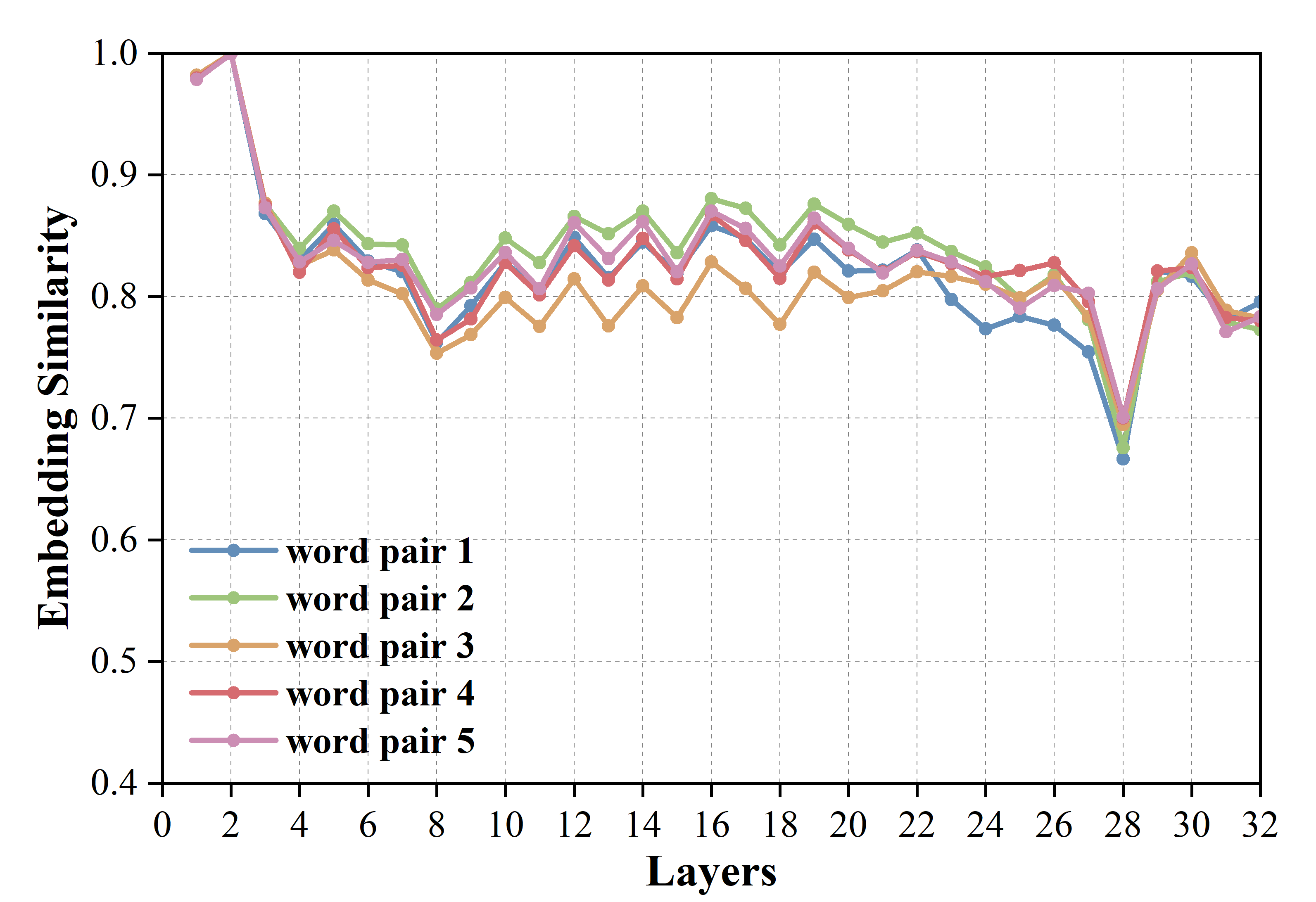}
\caption{
\textbf{Embedding semantic similarity} between Arabic-Hebrew 
and Chinese-Hebrew translations when predicting the same token
at each layer of LLaMA 3.
\\
}
\label{fig4:embeddingsim}
\end{figure}

Here we discuss
which $k$ middle layers should be selected 
to quantify linguistic similarity.
According to observations in section 
\ref{neuronovs},
neurons in middle layers
should be prioritized over final layers when measuring language similarity.
We use the embedding semantic similarity metric
to determine $K$.
For example,
we analyze Arabic-Hebrew and Chinese-Hebrew translation pairs
by prompting LLMs with the same semantical  words in Arabic and Chinese to generate the corresponding Hebrew translation.
We then compute the layer-wise embedding semantic similarity 
between the two pairs and 
identify layers in which this similarity is insensitive to variations in the predicted tokens,
reflecting the inherent distance between the languages.
As shown in Figure \ref{fig4:embeddingsim},
we find embedding similarity 
is stable in the middle layers 
10-21 of LLaMA 3 and 
is not affected by token-level variations.
Therefore,
$K$ layers is set to be 10-21 in LLaMA 3 
and 15-23 in Mistral.

\subsection{Overlap Neuron Distribution}
\label{sec:appendixond}
\begin{figure}[!h]
\centering
\includegraphics[width=0.45\textwidth]
{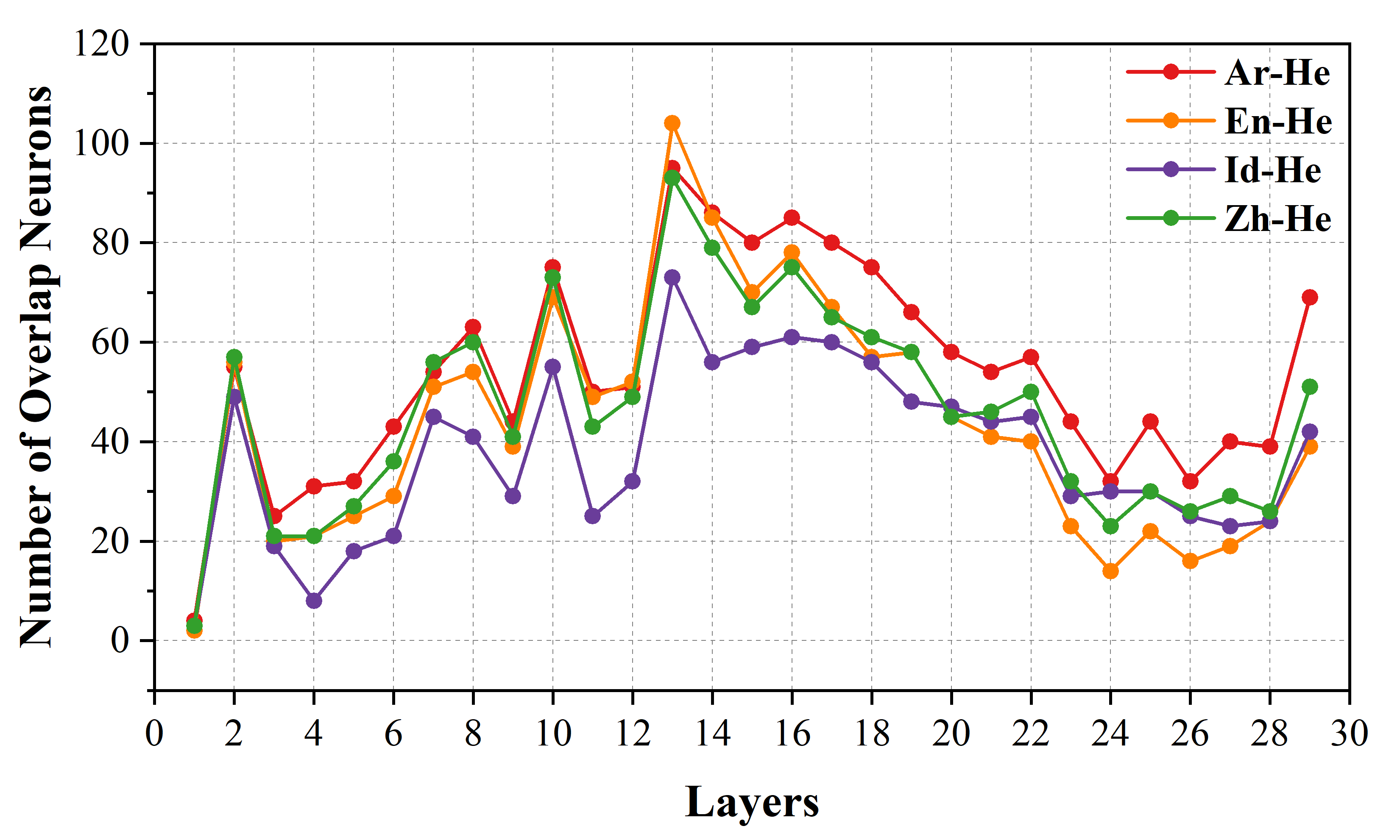}
\caption{\textbf{Distribution of 
overlap neurons} 
across different language pairs
in LLaMA 3.}
\label{fig:no_fresss}
\end{figure}

Figure \ref{fig:no_fresss}
presents the distribution of 
overlap neurons across different language pairs in LLaMA 3.

\section{Appendix: Ablation Results}
\label{sec:appendixBAlare}
\begin{figure}[!ht]
\centering
\includegraphics[width=0.45\textwidth]
{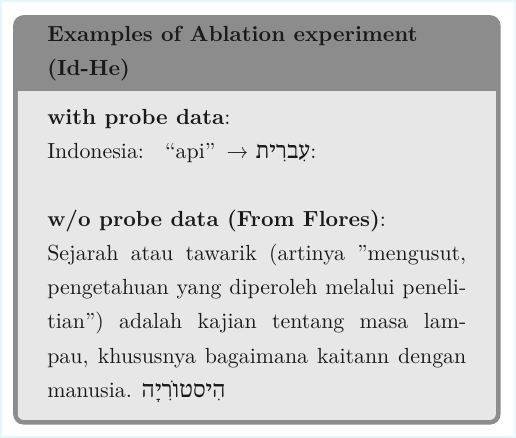}
\caption{\textbf{Example 
of ``w/o $*$'' probe data} in Indonesian-Hebrew.}
\label{fig:probedata*}
\end{figure}

This section presents the detailed experimental ablation to
evaluate the impact of neuron probe data construction and 
the HSIC similarity metric
on the bridge language selection. 
Table \ref{ablation-pro} 
and Table \ref{hsicaba} 
present the results of ablation experiments on the BLI task. ``w/o $*$'' denotes replacing our probe data with a simplified version based on FLORES+. 
For example,
Figure \ref{fig:probedata*}
illustrates the 
construction of 
``w/o $*$'' probe data for Indonesian-Hebrew.

\begin{table}[!h]
\caption{Performance comparison of using HSIC and Cosine similarity metrics on the BLI task.}
\centering
\resizebox{0.48\textwidth}{!}{
\scriptsize
\setlength{\abovetopsep}{1pt}
\setlength{\belowbottomsep}{1pt}
\setlength{\aboverulesep}{0pt}
\setlength{\belowrulesep}{0pt}
\begin{tabular}{l|cl|cl} 
\toprule[0.8pt]
 & \multicolumn{4}{c}{LLaMA-3-8B} \\ 
\cline{2-5}
 & \multicolumn{1}{l}{Bridge} & \multicolumn{1}{l|}{HSIC} & \multicolumn{1}{l}{Bridge} & \multicolumn{1}{l}{Cosine} \\
\hline
\textbf{Zh-Ja} & En & 77.90\,\textcolor{green}{$\uparrow$} & De & 76.60 \\
\textbf{Zh-He} & En & 56.70\,\textcolor{red}{$\downarrow$}& De & 58.60 \\
\textbf{Zh-Tl} & En & 54.00\,\textcolor{green}{$\uparrow$} & De & 51.00 \\
\textbf{Zh-Sw} & En & 36.00\,\textcolor{green}{$\uparrow$} & De & 33.80 \\
\textbf{Ar-Ja} & En & 80.60\,\textcolor{green}{$\uparrow$} & De & 76.30 \\
\textbf{Ar-He} & En & 64.50\,\textcolor{green}{$\uparrow$} & De & 64.10 \\
\textbf{Ar-Tl} & En & 56.80\,\textcolor{green}{$\uparrow$} & De & 52.80 \\
\textbf{Ar-Sw} & En & 42.40\,\textcolor{red}{$\downarrow$}& De & 43.00 \\
\textbf{Id-Ja} & En & 66.10\,\textcolor{green}{$\uparrow$} & De & 65.10 \\
\textbf{Id-He} & Es & 61.30\,\textcolor{green}{$\uparrow$} & De & 59.90 \\
\textbf{Id-Tl} & Fr & 61.00\,\textcolor{green}{$\uparrow$} & De & 57.00 \\
\textbf{Id-Sw} & Fr & 30.00\,\textcolor{green}{$\uparrow$} & De & 28.30 \\
\textbf{En-He} & Es & 71.80\,\textcolor{red}{$\downarrow$}& Pt & 72.30 \\
\textbf{En-Tl} & Fr & 58.70\,\textcolor{green}{$\uparrow$} & Pt & 55.80 \\
\textbf{En-Sw} & Fr & 26.20\,\textcolor{green}{$\uparrow$} & It & 21.50 \\
\bottomrule[0.8pt]
\end{tabular}
\label{hsicaba}
}
\end{table}

\section{Appendix: Neuron Semantic Analysis}

In this section, we analyze the semantic similarity of overlapping neurons across two language groups: Hebrew-Tagalog-Swahili (He-Tl-Sw, from different language families) and Portuguese-Spanish-Italian (Pt-Es-It, from the same language family). 
Our goal is to examine
whether overlapping neurons
cluster together 
within the same language
family or also across different families.

For comparison, we select $100$ overlapping neurons and 
randomly sample the same number 
of non-overlapping neurons for comparison. 
We feed the model with $m$ parallel sentences and then record the neuron activation frequency for each sentence,
obtaining three $m \times 100$ activation matrices for both overlapping and random neurons.
These matrices are mapped to a 2D semantic space using UMAP \cite{x13}, with each point representing a neuron activated by $m$ sentences from a language.
Colored circles of red, yellow, and blue denote overlapping neurons and colored $\times$ symbols of
red, yellow, and blue denote random neurons.

\begin{figure}[t]
  \centering
  \begin{minipage}[b]{0.45\textwidth} 
    \centering   
    \subcaption{Neuron semantic similarity in Pt-Es-It}
    \includegraphics[width=\linewidth, keepaspectratio]{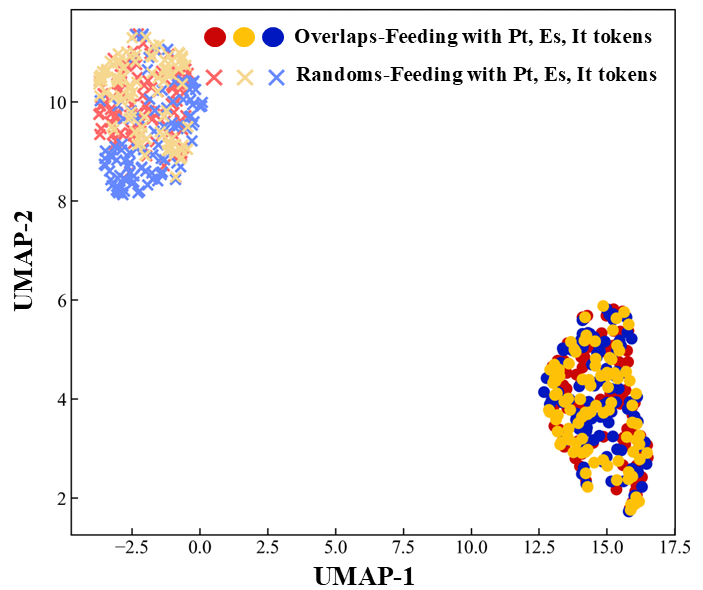} 
    \label{fig:subfigC}
  \end{minipage}
  \hfill 
  \begin{minipage}[b]{0.45\textwidth} 
    \centering
    \subcaption{Neuron semantic similarity in He-Tl-Sw}
    \includegraphics[width=\linewidth, keepaspectratio]{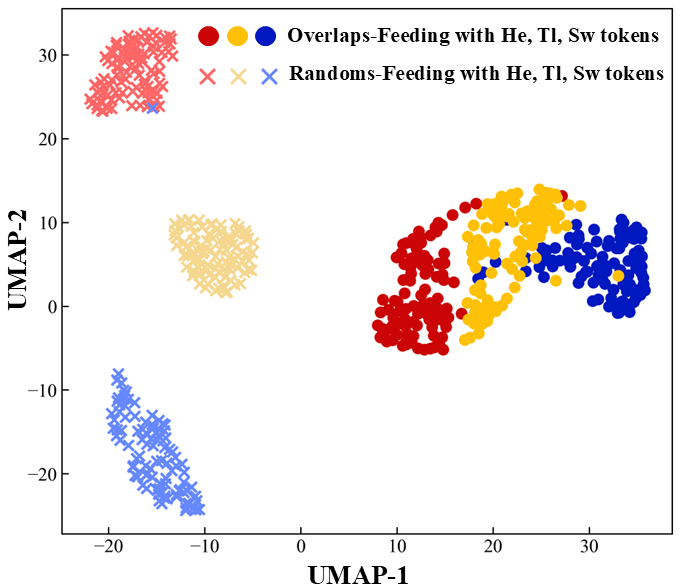} 
    \label{fig:subfigD}
  \end{minipage}

  \caption{\textbf{Visualization of semantic similarity} comparing language-overlapping neurons and randomly sampled neurons.} 
  \label{fig:semantic}
\end{figure}

As presented in Figure \ref{fig:semantic},
the overlapping neurons,
whether identified 
within the same family or across language families,
cluster closely together.
This suggests that our approach can effectively 
identify neurons that encode genuine
shared semantics across languages.
In addition, 
random neurons tend to align with linguistic relationships,
clustering when
feeding LLMs tokens from topologically related languages,
such as 
Portuguese (Pt),
Spanish (Es), and
Italian (It),
while remaining dispersed
when feeding LLMs tokens from distant
languages, 
like Hebrew (He),
Tagalog (Tl),
and Swahili (Sw).

\newpage
\section{Appendix: Prompt Templates}
\label{sec:appendixC_prompt}

{\setlength{\intextsep}{0pt}
 \setlength{\textfloatsep}{1pt}
 \begin{figure*}[]
 \centering
 \includegraphics[width=0.8\textwidth]{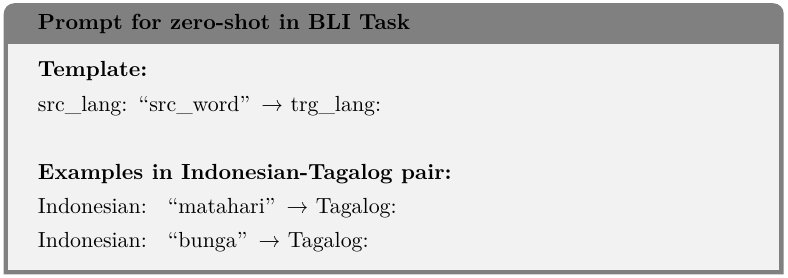}
 \end{figure*}}
 
{\setlength{\intextsep}{0pt}
 \setlength{\textfloatsep}{1pt}
 \begin{figure*}[]
 \centering
 \includegraphics[width=0.8\textwidth]{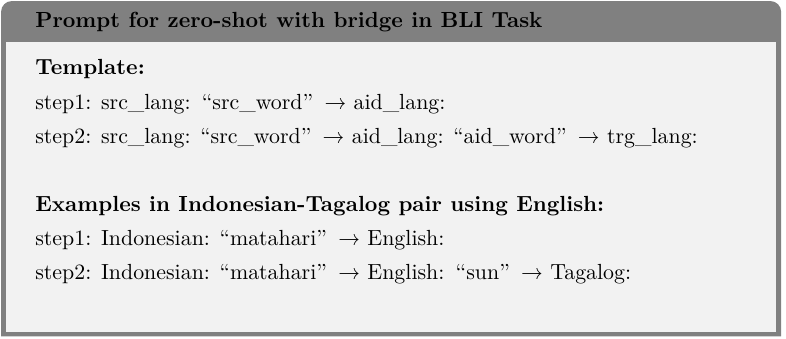}
 \end{figure*}}

{\setlength{\intextsep}{0pt}
 \setlength{\textfloatsep}{1pt}
 \begin{figure*}[]
 \centering
 \includegraphics[width=0.8\textwidth]{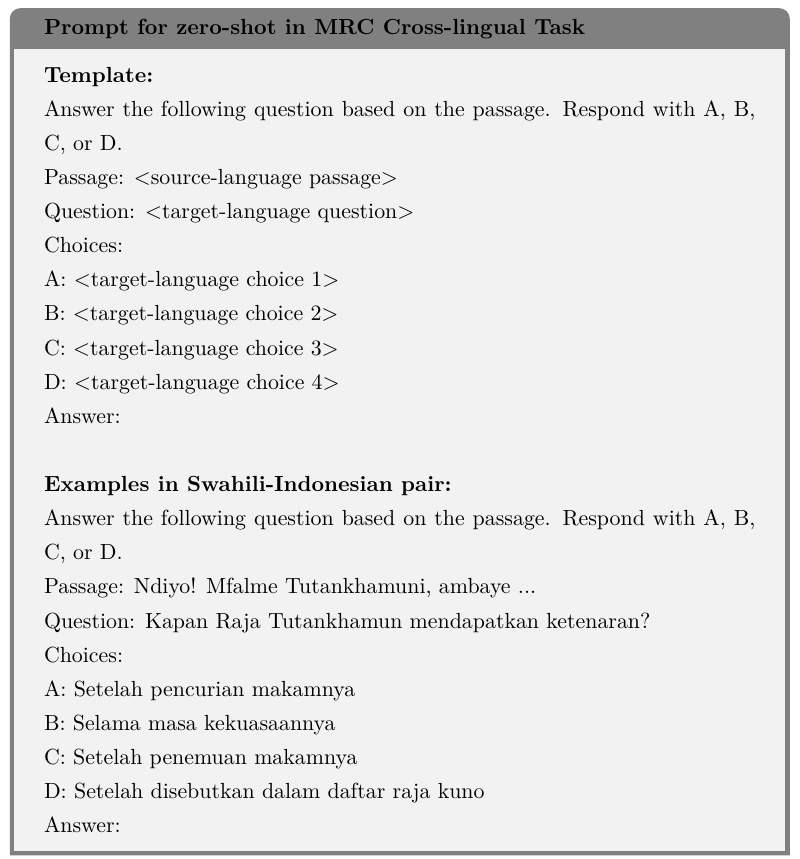}
 \end{figure*}}

{\setlength{\intextsep}{0pt}
 \setlength{\textfloatsep}{1pt}
 \begin{figure*}[]
 \centering
 \includegraphics[width=0.8\textwidth]{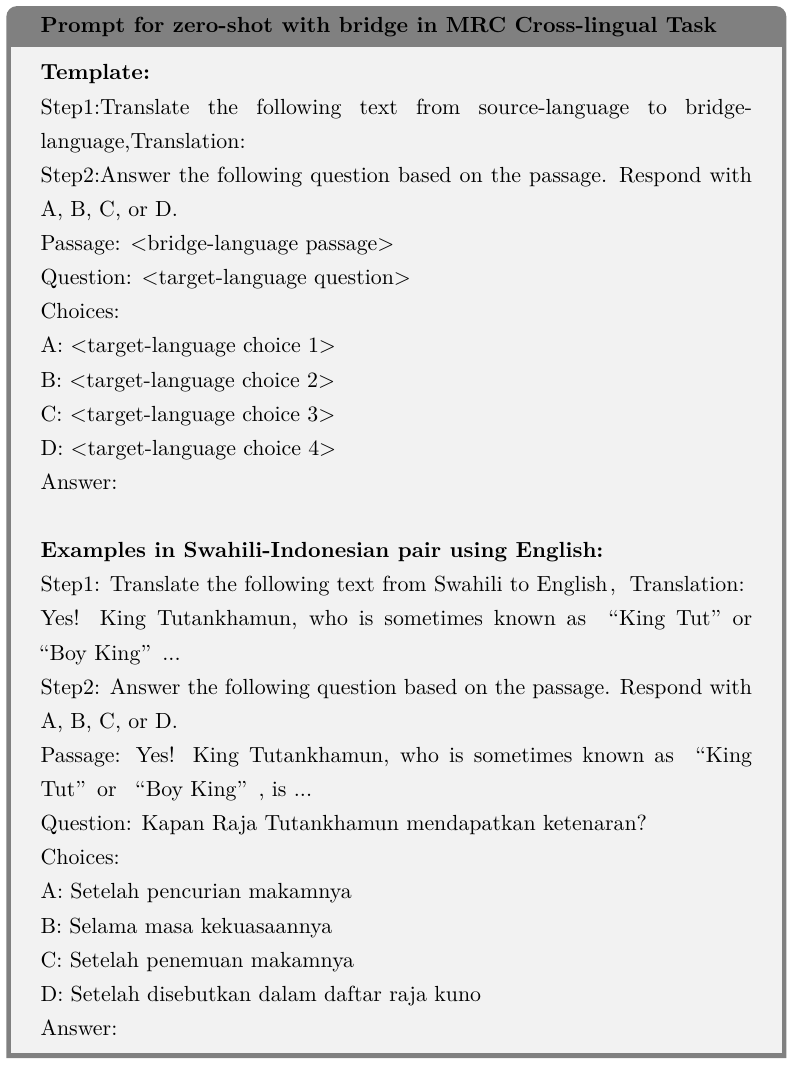}
 \end{figure*}}
 
\end{document}